\documentclass[lettersize,journal]{IEEEtran}
\usepackage{amsmath,amsfonts}
\usepackage{algorithmic}
\usepackage{algorithm}
\usepackage{array}
\usepackage[caption=false,font=normalsize,labelfont=sf,textfont=sf]{subfig}
\usepackage{textcomp}
\usepackage{stfloats}
\usepackage{url}
\usepackage{verbatim}
\usepackage{graphicx}
\usepackage{cite}
\graphicspath{{eps/}}
\usepackage{amssymb}
\usepackage{pifont} 

\begin{document}

\title{Federated Distillation: A Survey}

\author{ {Lin Li}, {Jianping Gou$^{\ast}$, \emph{Senior Member}, \emph{IEEE}}, {Baosheng Yu}, {Lan Du , \emph{Senior Member}, \emph{IEEE}}, {Zhang Yi, \emph{Fellow}, \emph{IEEE}}, {Dacheng Tao, \emph{Fellow}, \emph{IEEE}}
\thanks{Lin Li is with the School of Computer Science and Communication Engineering and Jiangsu Key Laboratory of Security Tech. for Industrial Cyberspace, Jiangsu University, Zhenjiang, 212013, P.R.China (linli@stmail.ujs.edu.cn) }
\thanks{Jianping Gou is with the College of Computer and Information Science, College of Software, Southwest University, Chongqing 400715, P.R.China (e-mail: cherish.gjp@gmail.com)}
\thanks{Baosheng Yu is with the School of Computer Science, Faculty of Engineering, The University of Sydney, Darlington, NSW 2008, Australia (e-mail: baosheng.yu@sydney.edu.au)}
\thanks{Lan Du is with the Faculty of Information Technology, Monash University, Clayton, VIC, Australia (e-mail: lan.du@monash.edu)}
\thanks{Zhang Yi is with the School of Computer Science, Sichuan University, Chengdu, 610065, P.R.China (Email: zhangyi@scu.edu.cn).}
\thanks{Dacheng Tao is with the School of Computer Science and Engineering, Nanyang Technological University, Singapore, 639798, Singapore (e-mail: dacheng.tao@gmail.com)}
\thanks{$^{\ast}$ Corresponding author.}}

\markboth{Journal of \LaTeX\ Class Files,~Vol.~14, No.~8, August~2021}%
{Shell \MakeLowercase{\textit{et al.}}: A Sample Article Using IEEEtran.cls for IEEE Journals}


\maketitle

\begin{abstract}
Federated Learning (FL) seeks to train a model collaboratively without sharing private training data from individual clients. Despite its promise, FL encounters challenges such as high communication costs for large-scale models and the necessity for uniform model architectures across all clients and the server. These challenges severely restrict the practical applications of FL. To address these limitations, the integration of knowledge distillation (KD) into FL has been proposed, forming what is known as Federated Distillation (FD). FD enables more flexible knowledge transfer between clients and the server, surpassing the mere sharing of model parameters. By eliminating the need for identical model architectures across clients and the server, FD mitigates the communication costs associated with training large-scale models. This paper aims to offer a comprehensive overview of FD, highlighting its latest advancements. It delves into the fundamental principles underlying the design of FD frameworks, delineates FD approaches for tackling various challenges, and provides insights into the diverse applications of FD across different scenarios.
\end{abstract}


\section{Introduction}
\label{sec1}

\IEEEPARstart{T}{he} success of deep learning can be largely attributed to the abundance of available training data, as evident in studies like  Krizhevsky et al. \cite{krizhevsky2017imagenet} and AlphaGo \cite{silver2016mastering}. However, applying deep learning in real-world scenarios encounters significant challenges due to industry-specific traits and legal regulations. Firstly, training data is often scattered across diverse, isolated devices, posing a challenge in consolidating it for model training. Secondly, the growing emphasis on data privacy and security requires safeguarding locally sensitive data. In response to these challenges, federated learning (FL) has been introduced. FL enables collaborative model training without the need to share private training data from individual clients. Despite its potential benefits, FL faces obstacles such as high communication costs for large-scale models and the necessity for all clients to adopt the same model architecture as the server.

\begin{figure}[!ht]
\centering
\includegraphics[width=3.5in]{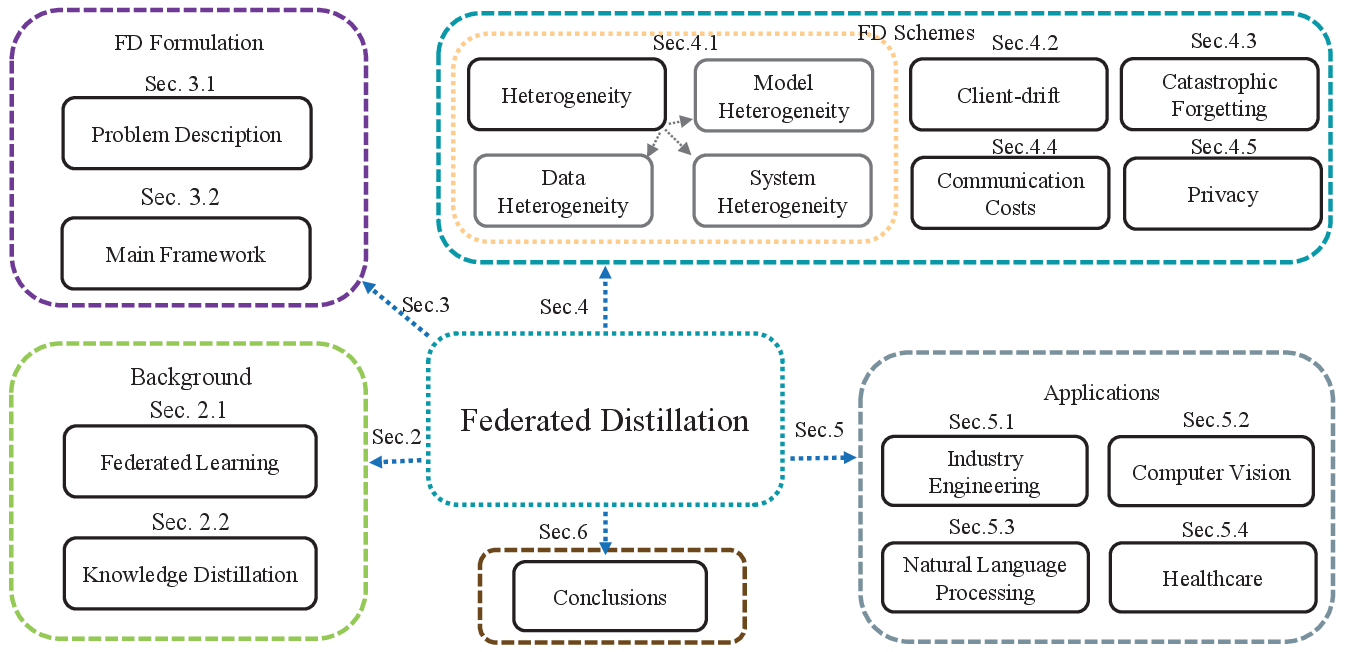}
\caption{An overview of the organization of the different sections in this paper.}
\label{fig1}
\end{figure}

Initially, the integration of knowledge distillation (KD) into FL aimed to enhance communication between clients and the server~\cite{jeong2018communication}. Subsequently, Li et al. \cite{li2019fedmd}, Sattler et al. \cite{sattler2020communication}, and Seo et al. \cite{seo202216} explored KD for FL methods, termed Federated Distillation (FD). These approaches primarily focus on client-side model training, leveraging KD to compress the FL backbone network and improve the global model's performance through knowledge transfer. While they offer partial data privacy protection, challenges persist, including minimizing the information shared with the teacher model and addressing dataset heterogeneity. The rapid advancement of FD is reflected in the increasing number of research papers published on the topic~\cite{itahara2021distillation, han2022fedx, kim2022multi, wen2022communication}.

Current FD methods primarily focus on enhancing performance by transferring intermediate features or logits from the teacher to the student model~\cite{ cheng2021fedgems, mishra2021network, zhang2021parameterized}. Numerous innovative modules or mechanisms have been proposed to enhance FD performance, including synthetic sampling \cite{zhang2023towards}, bidirectional co-distillation \cite{ni2022federated, qi2022fedbkd, shang2023fedbikd}, two-way mixup \cite{oh2020mix2fld}, reinforcement distillation \cite{cha2019federated,cha2020proxy}, mutual distillation \cite{li2023fedh2l}, differential privacy \cite{li2021practical}, and meta-algorithm \cite{taya2022decentralized}. Furthermore, various efforts have been made to apply FD in different applications such as wireless communications \cite{wen2020unified}, keyword spotting \cite{hard2022production}, person re-identification \cite{zhuang2020performance}, intelligent networked vehicle systems \cite{zhang2021distributed}, intrusion detection \cite{zhao2022semi}, and classification \cite{lin2020ensemble}.

With the increasing interest and successful applications of FD in various fields~\cite{wang2020industrial}, there arises a pressing need for a comprehensive survey on FD. Notably, this paper marks the first systematic review of KD strategies applied to FL. Existing reviews on KD predominantly focus on summarizing strategies and application scenarios, yet they lack a specific examination of FL \cite{gou2021knowledge, li2023object, tian2023knowledge}. Conversely, while reviews of FL address its mechanisms and challenges, they fail to adequately consider KD models \cite{li2020review, mothukuri2021survey, zhu2021federated, zhang2022challenges, cao2023bayesian, li2020federated, yang2019federated, yang2022survey}. Although some reviews on FD do exist \cite{mora2022knowledge, wu2023survey}, they primarily concentrate on summarizing KD applications in FL, without delving into a detailed examination of the challenges faced or addressed by FL. Therefore, this survey aims to fulfill three main goals.
Firstly, it offers a systematic review of FD literature, explaining how KD addresses FL challenges. It provides a comprehensive summary of existing methods, serving as a valuable reference for KD and FL researchers. Secondly, it identifies key research challenges and directions, summarizing coping strategies and suggesting areas for future FD research.
Lastly, the survey provides an overview of popular FD applications.

As illustrated in Fig.~\ref{fig1}, the following sections are organized as follows: In Section 2, we introduce the concepts of FL and KD. Section 3 thoroughly explains the main FD formulation. To address different FL challenges, Section 4 summarizes various FD schemes for tackling specific challenges. Section 5 explores the applications of FD across various domains and suggesting potential avenues for future research. Finally, Section 6 concludes the paper by summarizing key findings.

\section{Background}
\label{sec2}

\subsection{Federated Learning}
\label{sec2.1}

In recent years, deep learning algorithms have advanced significantly, powered by vast amounts of training data from mobile devices like smartphones and self-driving cars. This data abundance has propelled the rapid growth of artificial intelligence. However, privacy concerns have become a crucial issue. People are increasingly hesitant to share personal information, especially sensitive data like medical records and shopping habits, due to the awareness of potential privacy breaches \cite{gong2020privacy, albrecht2016gdpr, parasol2018impact}.

Governments worldwide have acknowledged the need to protect personal data and have implemented laws and regulations to ensure individuals' privacy. For instance, in China, the Cyber Security Law of the People's Republic of China \cite{parasol2018impact} and the General Principles of the Civil Law of the People's Republic of China \cite{gray1986general} were enacted in 2017. These regulations prohibit network operators from tampering with, destroying, or disclosing clients' personal information. In Europe, the General Data Protection Regulation (GDPR) \cite{albrecht2016gdpr} came into effect on May 25th, 2018. The GDPR aims to safeguard the privacy and security of users' personal data, requiring operators to obtain clear user consent and prohibiting them from inducing or deceiving users into waiving their privacy rights. Additionally, the GDPR grants users the right to request the deletion of their personal information.

In today's setup, data storage is increasingly organized into data silos, where data from multiple devices is kept separate to ensure client data's privacy and security. While this safeguards user privacy, it also poses significant challenges in terms of data availability. Gathering enough data for training deep learning models becomes harder due to data silos. Consequently, the traditional deep learning framework is no longer suitable in the context of data silos. Finding a solution to overcome the constraints imposed by data silos while safeguarding data privacy has become a widely discussed and explored topic in the field of AI. There's an urgent need for a new modeling framework that can effectively tackle the challenges posed by data silos while preserving users' data privacy.

\begin{figure}[!t]
\centering
\subfloat[]{\includegraphics[width=3.5in]{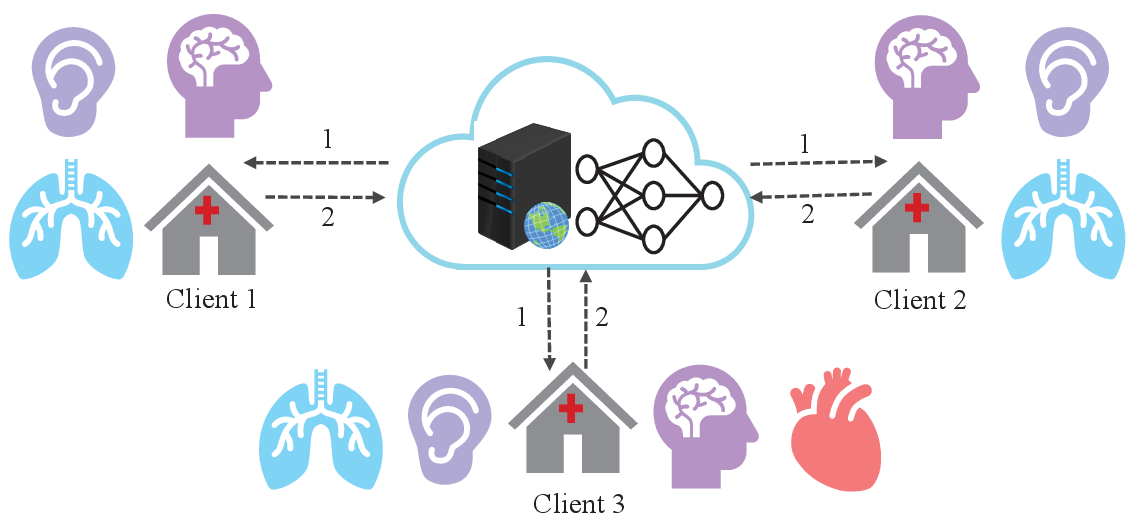} \label{fig2.1}}
\hfil
\subfloat[]{\includegraphics[width=3.5in]{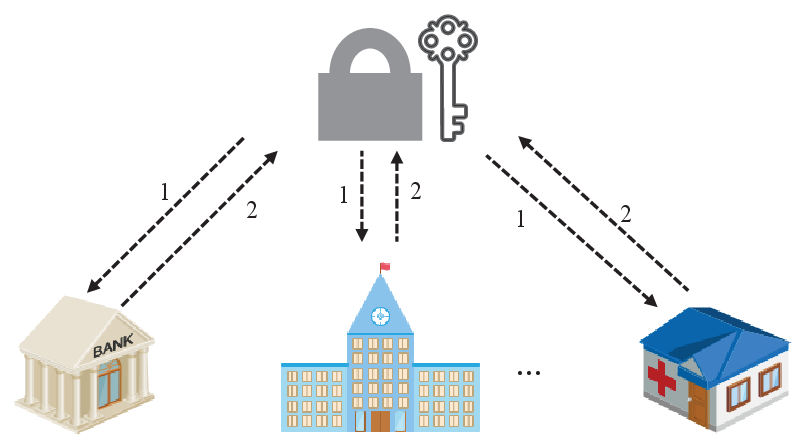}\label{fig2.2}}
\hfil
\subfloat[]{\includegraphics[width=3.5in]{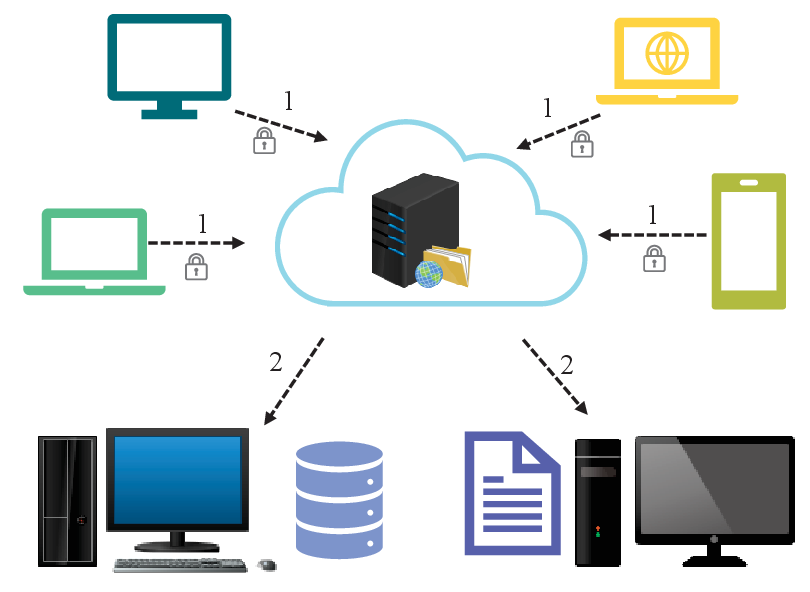} \label{fig2.3}}
\caption{An overview of three main federated learning categories. (a) Horizontal Federated Learning. Step 1: Download the trained global model and repeat the training cycle for use on each client node. Step 2: Homogeneous clients from the same domain for training the global model utilizing private data. (b) Vertical Federated Learning. Step 1: Download the trained global model and repeat the training cycle for use on each client node. Step 2: Heterogeneous clients assist in training the global model by sharing encrypted local model updates. (c) Federated Transfer Learning. Step 1: Train the global model with a heterogeneous client in the encrypted state, similar to VFL. Step 2: Obtaining personalized local models from global model through transfer learning.}
\label{fig2}
\end{figure}

In 2017, Google introduced a groundbreaking framework called federated learning, revolutionizing how shared models are trained in a decentralized manner \cite{mcmahan2017communication}. This framework enables the collaborative training of a model without clients having to upload their private datasets to a central server. Instead, each client performs computations locally, generating an update information to the global model, which is then communicated solely between the client and the server. This approach ensures that individuals can benefit from a well-trained AI model without compromising the privacy of their personal information. Since then, FL has emerged as a promising deep learning framework, driven by the growing concerns surrounding the privacy of personal data.
Current FL methods can be grouped into three main types based on the characteristics of data distribution \cite{yang2019federated,nguyen2021federated} as follows:\\
\begin{enumerate}[]
    \item[1)]
{\bf Horizontal Federated Learning (HFL)} is suitable when there are few overlapping users in two datasets but more overlapping user features \cite{zhu2021federatedsurvey}, as shown in Fig.~\ref{fig2.1}. In this scenario, the datasets are horizontally divided, and data with the same features but from different users are chosen for training. For instance, consider two hospitals located in different regions. Although there is a small overlap of users between them, they offer the same type of services and record similar types of data. Each hospital possesses a private dataset containing information about the same type of disease. By utilizing horizontal FL, these hospitals can collaborate to train an improved disease diagnosis model.\\

\item[2)] {\bf Vertical Federated Learning (VFL)} is useful when two datasets share many users but have fewer overlapping user features \cite{yang2023survey,zhang2021survey}. The goal is to train a more comprehensive model by combining different data features from users. For example, consider a bank and a hospital, both serving the local population, operate in the same area. The bank tracks user income, expenditure, and credit history, while the hospital maintains patient health records. Consequently, the overlap of user features between these two institutions is relatively small. VFL merges these distinct features in the encrypted state, and Fig.~\ref{fig2.2} provides an illustration of VFL. \\

\item[3)]{\bf Federated Transfer Learning (FTL)} comes into play when both datasets have limited overlap in features and samples, which utilizes transfer learning to overcome the lack of data or labels \cite{li2020review}, as shown in Fig.~\ref{fig2.3}. For example, consider hospitals and banks situated in different regions. Due to geographical constraints, there is minimal overlap of users between these two organizations. Additionally, the difference in institution type results in only a small overlap in their data characteristics. In such cases, FTL can be employed to tackle the challenge of limited samples available for client data, thereby improving the model's performance.\\
\end{enumerate}

\begin{figure}[!t]
\centering
  {\includegraphics[width=3.5in]{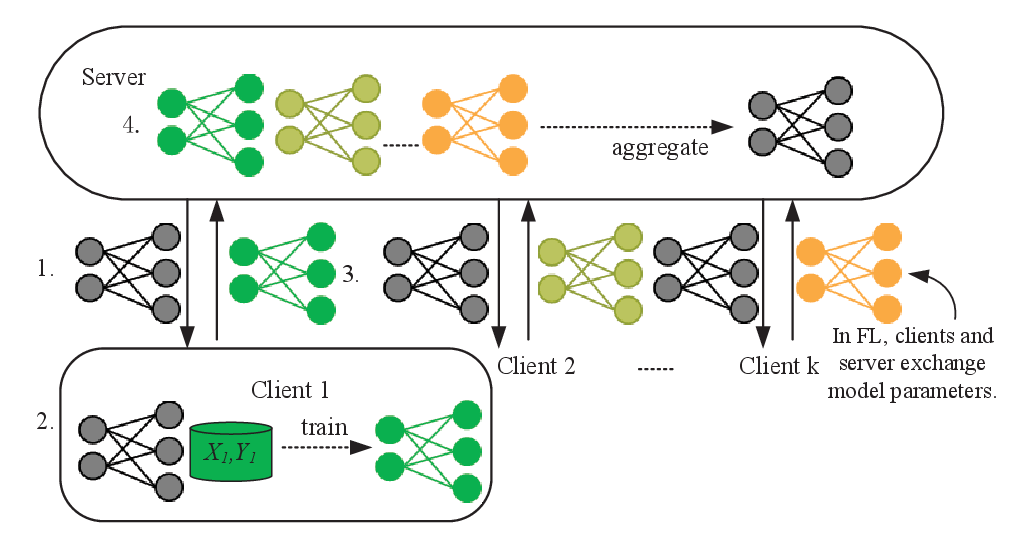}}
  \caption{Illustration of the vanilla federated averaging framework.}
  \label{fig3}
\end{figure}

FL has seen significant progress in privacy-preserving algorithms, giving rise to several notable ones like FedAvg \cite{mcmahan2017communication}, FedProx \cite{li2020federated}, FedBN \cite{li2021fedbn}, and MOON \cite{li2021model}. Among these, FedAvg is a classic algorithmic approach to FL, illustrated in Fig.~\ref{fig3}. It involves a central server and $K$ clients, each with their private local dataset. The process of a single communication round in FedAvg can be summarized as follows:
\begin{enumerate}[]
    \item The server broadcasts the current model parameters to a subset selected from $K$ clients for local model setup at the beginning of each training iteration.
    \item The chosen local client uses the initialized model to train on its private data, enhancing the model's performance.
    \item The client sends back the updated model (i.e., the model parameters after training with the local dataset) to the server.
    \item The server combines the model update parameters to create a new global model, serving as the client's starting model for the next training round.\\
\end{enumerate}

In regular FL, the exchange of information between local and global models only involves model parameters. Various surveys provide detailed descriptions and classifications of FL approaches \cite{li2020review, mothukuri2021survey, zhu2021federatedsurvey, zhang2022challenges, cao2023bayesian, li2020federated}.
While FL is commonly used for intelligent models with decentralized data, frameworks like parameter-averaged aggregation FL (e.g., FedAvg) have clear drawbacks. These algorithms require models within the framework to be homogeneous, demanding client models to have the same network architecture. The process of exchanging or updating parameters in the models incurs high communication costs due to larger models, and massive parameter updates or exchanges increase the risk of information leakage. In cases of client data heterogeneity, local parameters of the client model may deviate from the global optimum of the server model, leading to the client drift phenomenon. The introduction of KD effectively addresses these issues.

\subsection{Knowledge Distillation}
 \label{sec2.2}
Deep learning has introduced various architectures excelling in solving diverse problems. However, their application in resource-limited platforms, characterized by memory constraints and limited high-performance computing power, and in scenarios requiring fast execution, is restricted due to their large number of model parameters. Recently, KD has emerged as a highly successful approach for robust and efficient model compression. It enables the compression of well-trained, large networks into compact models suitable for deployment on various devices. KD was initially introduced by Hinton et al. \cite{hinton2015distilling} with the aim of distilling knowledge acquired from multiple models or a complex model into a simpler student model, intending to maintain performance while reducing complexity.

\begin{figure}[!ht]
\centering
  {\includegraphics[width=3.3in]{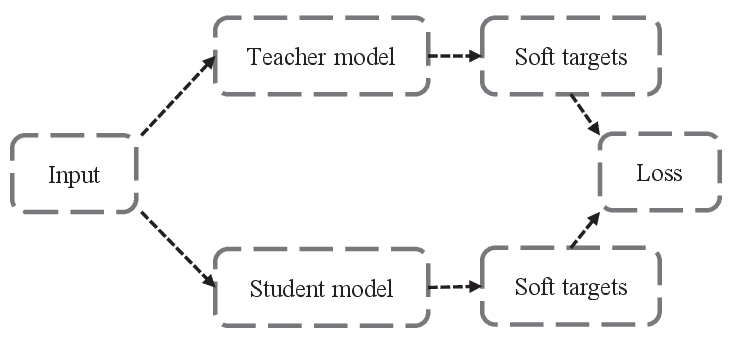}}
  \caption{The classical KD framework.}
  \label{fig4}
\end{figure}

The groundbreaking work of \cite{hinton2015distilling} introduced the use of soft labels, derived from the teacher model's output, to guide the learning process of the student model as follows:
\begin{equation}\label{eq1}
L_{soft}=L_{KL}\left ( p\left ( z_{t},T  \right ) ,p\left ( z_{s},T  \right ) \right ),
\end{equation}
where $L_{KL}$ is the Kullback-Leiber divergence loss, $T$ represents the temperature parameter, and $p\left ( z_{t},T  \right )$ and $p\left ( z_{s},T  \right )$ are soft labels probabilities can be estimated by a softmax function of teacher and student, respectively.
\begin{equation}
\small
{p\left ( z_{t},T  \right )} = \frac{{\exp \left( {{{{z_{ti}}} \mathord{\left/
 {\vphantom {{{z_{ti}}} T}} \right.
 \kern-\nulldelimiterspace} T}} \right)}}{{\sum\nolimits_{i=1}^N {\exp \left( {{{{z_{ti}}} \mathord{\left/
 {\vphantom {{{z_{ti}}} T}} \right.
 \kern-\nulldelimiterspace} T}} \right)} }},
 \quad
 {p\left ( z_{s},T  \right )} = \frac{{\exp \left( {{{{z_{si}}} \mathord{\left/
 {\vphantom {{{z_{si}}} T}} \right.
 \kern-\nulldelimiterspace} T}} \right)}}{{\sum\nolimits_{i=1}^N {\exp \left( {{{{z_{si}}} \mathord{\left/
 {\vphantom {{{z_{si}}} T}} \right.
 \kern-\nulldelimiterspace} T}} \right)} }},
 \end{equation}
wherein $N$ represents the total number of classes in the dataset, $z_{ti}$ and $z_{si}$ denote the logits from teacher model $z_t$ and student model $z_s$ on the $i$-th class, respectively. The overall framework is illustrated in Fig.~\ref{fig4}. However, relying solely on soft targets for learning is insufficient in creating student models with the desired performance, especially as the teacher model became challenging.

To overcome this, the concept of knowledge was expanded to include various aspects, such as hint layer features, output layer responses, inter-layer relationships, and any other relevant information beneficial for knowledge transfer \cite{gou2021knowledge}. Existing KD methods fall into four main categories based on the type of knowledge they utilize, as follows:\begin{enumerate}[]
\item {\it Label Knowledge} --- This category of KD involves utilizing the final output, which includes valuable information from the predictions of neural networks on sample data. The total distillation loss can be written as
\begin{equation}\label{eq2}
{L_{total}} =  {L_{soft}} + \alpha {L_{hard}}\,,
\end{equation}
where $\alpha$ is the equilibrium constant, controlling the trade-off between two losses. $L_{hard}$ involves calculating cross-entropy for the probability distribution of the student model based on label knowledge. On the other hand, $L_{soft}$ involves calculating cross-entropy when the probability distributions of both the student model and the teacher model are processed with a temperature parameter.
Several techniques have been proposed to enhance label KD. For example, Wen et al. introduces a logits adjustment mechanism to reduce the impact of incorrect supervision \cite{wen2021preparing}. ESKD suggests using an early stopping mechanism for teachers to facilitate effective student learning from teacher logits \cite{cho2019efficacy}. DKD divides the logits information using different views, exploring the significant potential of logits information in guiding student network training \cite{zhao2022decoupled}. These advancements aim to improve the quality and effectiveness of knowledge transfer from the teacher model to the student model. \\

\item {\it Feature Knowledge} ---  In this category of KD, intermediate-hidden features extracted from the middle layers of the network are utilized as knowledge. The feature knowledge extracted from the teacher model serves as a direct guide for training the student model \cite{guan2020differentiable}. The middle-layer distillation loss is defined as:
\begin{equation}\label{eq3}
	{L_{feature}} = {\left\| {{f_t}{\left( x \right)} - {f_s}{\left( x \right)}} \right\|_n}\,,
\end{equation}
where $f_t$ and $f_s$ represent the intermediate layer features of the teacher model and the student model, respectively. $n$ represents the norm value, and $x$ denotes the sample. Specifically, FitNets was among the first methods to use intermediate features from the teacher model. It guides the student's intermediate features to resemble those of the teacher and predict similar outputs \cite{adriana2015fitnets}. FEED enables multiple teacher networks to distill knowledge at the feature map level through non-linear transformations, aiding knowledge transfer to the student model \cite{park2019feed}. Another approach by DKKR \cite{chen2021distilling} is a KD review strategy that uses information about the teacher's multi-level features to teach the student network layer by layer. These methods focus on transferring knowledge from the middle layers of the teacher model to guide the training of the student model, helping the student model learn crucial representations and features at an intermediate level. \\

\item {\it Parameter Knowledge} --- In this category, partially trained parameters or network modules of the teacher model are directly used as knowledge during distillation training. This method is often combined with other distillation approaches to enhance knowledge transfer \cite{chen2021local}. Specifically, PESF-KD proposes an efficient approach to transferring knowledge from the teacher network to the student network by updating specific parameters of the pre-trained teacher model \cite{rao2022parameter}. FSKD compresses the pre-trained teacher model to obtain a student model and adjusts the feature dimensions using $1\times1$ convolutions between layers, allowing the student model to achieve performance comparable to traditional fine-tuned distillation methods with a small number of samples \cite{li2020few}.
Similarly, IAKD \cite{fu2021interactive} and NGFSKD \cite{shen2021progressive} assist distillation by replacing modules of the teacher model and the student model. The distillation loss for individual module replacement is expressed as:
\begin{equation}\label{eq4}
    {L_{model}} = {L_{CE}}\left( {M_o^T{\left( x \right)} - M_o^S{\left( x \right)}} \right)\,,
\end{equation}
where $L_{CE}$ represents the calculation of cross-entropy, and $M_o^T$ and $M_o^S$ are the outputs of the corresponding modules in the teacher and student networks.
Recently, SAKD proposes merging teacher and student networks of the same style system into a multi-path network. During training, a different teacher network module is dynamically selected to replace the corresponding student network module for each sample, resulting in a student model with superior performance \cite{song2022spot}. These methods leverage parameter knowledge from the teacher model to guide the training of the student model, either by updating specific parameters or by replacing modules. This enables the student model to learn from the partially trained teacher model, improving its performance and reducing its complexity. \\

\item {\it Relation Knowledge} --- This category of KD methods utilizes relationships between samples (inter-class relationships) or contextual relationships within sample features (intra-class relationships), complex relationships between samples represented by a graph, and other structural information as knowledge. These relationships can exist at various stages of the model. A single structural relationship distillation loss can be expressed as:
     \begin{equation}\label{eq5}
        {L_{relation}} = F\left( {R_t{\left( x \right)},R_s{\left( x \right)}} \right)~,
     \end{equation}
    where $F$ represents the structural relation loss function, and $R_t$ and $R_s$ denote the relation information from the corresponding layer in the teacher and student networks. The location of the knowledge within the model can vary and may be at the output layer class or intermediate layer features \cite{hou2020inter}. Specifically,
    RKD proposes using distances between samples and angle relationships within sample features for distillation \cite{park2019relational}. CCKD \cite{peng2019correlation} and SPK \cite{tung2019similarity} exploit the similarity of sample characteristics in the teacher and student networks for knowledge transfer, while CRD introduces a contrastive learning strategy to better extract knowledge of similarities between features \cite{tian2020contrastive}. To achieve superior performance, several methods of mixture distillation for different structural features have been proposed, taking advantage of image characteristics \cite{liu2019structured,liu2020structured,de2022structural,zhang2022pointdistiller,liu2022multi,shu2021channel,lin2023simple}. These structural KD approaches mainly utilize pair-wise relationships between samples or within sample features to transfer knowledge from the teacher model to the student model. By incorporating these structural relationships, the student model can benefit from the information encoded in the teacher's model and improve its performance.\\
\end{enumerate}

Initially, KD was introduced as a method to compress large models and transfer their knowledge to smaller models. In recent years, KD has evolved beyond model compression, thanks to its various mechanisms that allow for tailored knowledge transfer to meet the needs of different models. It has proven successful in applications like visual recognition, natural language processing, and speech recognition, as shown in a recent review \cite{gou2021knowledge}. This broadening of KD's applicability underscores its versatility and effectiveness in transferring knowledge across different domains and tasks, paving the way for the development of more efficient and accurate models.

\section{FD Formulation}
\label{sec3}

KD is introduced to the FL framework to exchange information between the client and the server through logits. This not only improves data privacy but also significantly reduces the amount of information transmitted during communication, thus reducing communication costs. The superior performance of KD also helps mitigate the impact of client model architecture and data heterogeneity in FL. This approach led to the development of the FD framework, allowing clients to benefit from personalized machine learning models while securely sharing acquired knowledge with each other \cite{li2019fedmd}. The FD framework ensures privacy is maintained during the knowledge transfer process, making FL a powerful approach for collaborative learning across distributed systems.

The literature review indicated that KD has been incorporated into various FL frameworks, with only a few algorithms addressing federated transfer learning \cite{afonin2022towards,chen2020fedhealth} and vertical FL \cite{huang2023vertical}. The current algorithms introducing KD to FL primarily address challenges related to data-decentralized scenarios where samples exhibit similar characteristics. In other words, these methods use horizontal FL frameworks to introduce KD and alleviate corresponding challenges. Therefore, the subsequent section describes its typical FD framework.

\begin{figure}[!ht]
\centering
  \includegraphics[width=3.5in]{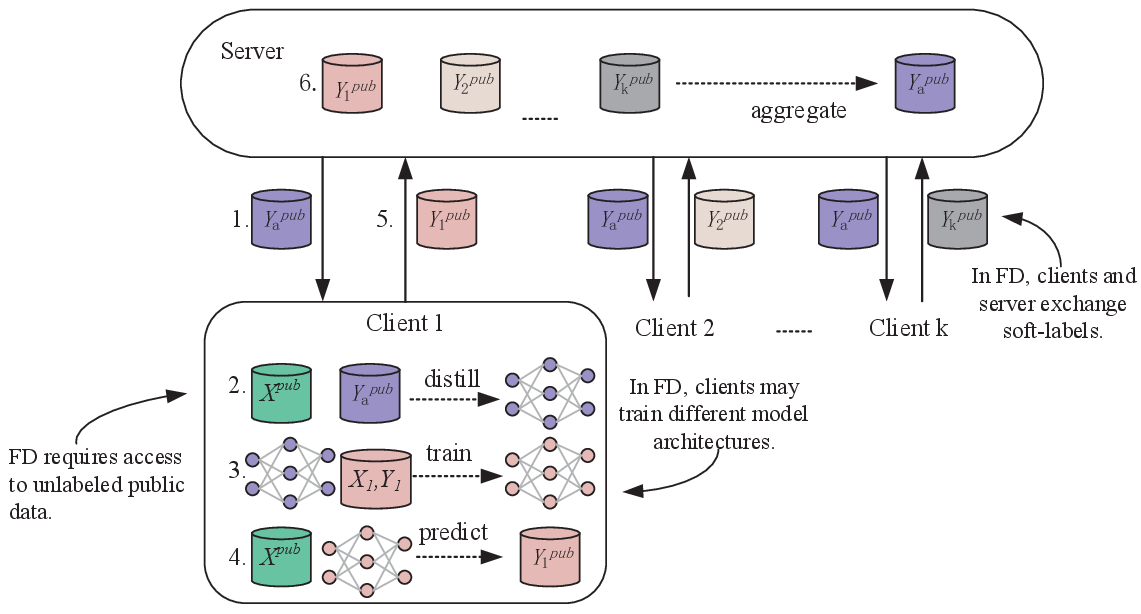}
  \caption{The main FD framework.}
  \label{fig5}
\end{figure}

\subsection{Problem Description}
 \label{sec3.1}

We introduce FD within the context of a standard horizontal FL framework, i.e., FedAvg, as follows. Consider a group of $K$ clients, each with its private dataset ${D_k} = \left\{ \left( {X_k, Y_k} \right) \right\}$, where $k \in \left\{ 1, 2, \ldots, K \right\}$. In each dataset $D_k$, there are $N_k$ training samples denoted as $X_k = \left\{ x_{kj} \right\}_{j=1}^{N_k}$ with corresponding labels $Y_k = \left\{ y_{kj} \right\}_{j=1}^{N_k}$, where $y_{kj} \in \left\{ 1, 2, \ldots, M \right\}$ and $M$ denotes the sample category. It's important to note that datasets $D_k$ may come from different distributions. Let $C$ represent a subset of clients, where $1 \leq C \leq K$. Also, let $X^{pub}$ be a public dataset accessible to all clients. The server model $\theta$ aggregates soft labels from the clients, denoted as $Y_a^{pub}$, while the client networks produce soft labels on the public dataset, represented as $Y_k^{pub}$.

During client training using private data, each client establishes its own local model $\theta_{k}$, and the corresponding model output is denoted as $f\left(\theta_{k} \right)$. With the loss function $\mathcal{L}(\cdot, \cdot)$, the empirical loss associated with client $k$ is defined as:
\begin{eqnarray}\label{eq6}
     \mathcal{L}(f\left( \theta_{k}\right), Y_{k} ) \triangleq \frac{1}{|{D_k}|} \sum_{j=1}^{N_k} \mathcal{L}\left(f(\theta_{k}), y_{kj}\right)\,.
\end{eqnarray}
In each communication round, each individual client optimises $\theta_{k}$  using its private data by minimizing the above loss function as
\begin{equation}\label{eq7}
     \theta^*_{k} = \underset{\theta_{k}}{argmin} \mathcal{L}(f\left( \theta_{k}\right), Y_{k} )\,.
\end{equation}
This implies that each client contributes improvements to the current model using its unique private data. The server then performs an aggregation operation for uploaded information from these refined models. In FD, the process is to optimize the information exchanged between the client and the server through KD, with the aim of minimizing the risk to data security.

\subsection{Main Framework}
 \label{sec3.2}

FD introduces a unique method for sharing knowledge acquired from training on local datasets, involving both homogeneous networks and heterogeneous model structures. The main FD framework is illustrated in Fig.~\ref{fig5}. Instead of directly sending locally trained model parameters to the server, FD communicates soft-label predictions $Y^{Pub}_k$ generated by each client to a publicly accessible distillation dataset $X^{Pub}$. In this manner, $Y^{Pub}_k$ acts as a concise form of knowledge transfer  between each client and the server within the distillation context.

In FD, ensuring a similar distribution between the public dataset and clients' private datasets is crucial, allowing the public dataset to serve as the distilled data source. In each communication round of the FD framework, participating clients retrieve aggregated soft label information $Y_a^{pub}$ from the server. This information is used to update clients' local models through distillation. To maintain consistency in the distillation process across clients, a random seed is introduced to manage inherent randomness, ensuring that all clients converge to the same distillation model. Subsequently, the updated local model is fine-tuned by training it with the client's private data, resulting in an optimized local model on each client. The client's network transmits the soft label $Y_k^{Pub}$ associated with predictions for the public data to the server. The server employs aggregation procedures to prepare for the subsequent round of communication. A detailed description of the steps involved in a communication round is provided below:
\begin{enumerate}
    \item A subset of $C$ clients downloads the soft labels $Y_a^{pub}$ from the server. The soft labels is associated with the public dataset ${X^{pub}}$.
    \item  The participating client uses the downloaded soft label information $Y_a^{pub}$ for model distillation, updating its local model $\theta_{k}$ based on the public dataset.
    \item After distillation, each client fine-tunes the model using its private data to enhance performance. This results in a personalized and optimized model for each client.
    \item The client retrains its locally optimized model on the public dataset ${X^{pub}}$ and then sends the predicted soft labels $Y_k^{pub}$ to the server.
    \item The server aggregates the soft labels in preparation for the next round of communication training.

\end{enumerate}

FD presents several advantages over traditional FL. By replacing traditional model parameters with predicted soft labels, FD effectively overcomes communication bandwidth limitations. This allows clients to diversify their model structures, catering to the constraints of hardware-limited devices. Moreover, FD enhances model robustness. Unlike conventional FL, which communicates via model parameters and may inadvertently reveal original data, FD's use of logits information makes compromising private data more challenging. This is supported by studies such as those by Faug \cite{jeong2018communication}, DS-FL \cite{itahara2021distillation}, FedDF \cite{lin2020ensemble}, and CFD \cite{sattler2021cfd}. Additionally, the incorporation of KD can further improve model performance \cite{liu2022efficient,bhatt2023federated, kim2023protofl,xie2023perada}.

\section{FD Schemes}
\label{sec4}

This section explores FD schemes designed to tackle fundamental challenges, including issues related to heterogeneity, communication costs, and privacy concerns.

 \begin{figure}[!ht]
\centering
  {\includegraphics[width=3.5in]{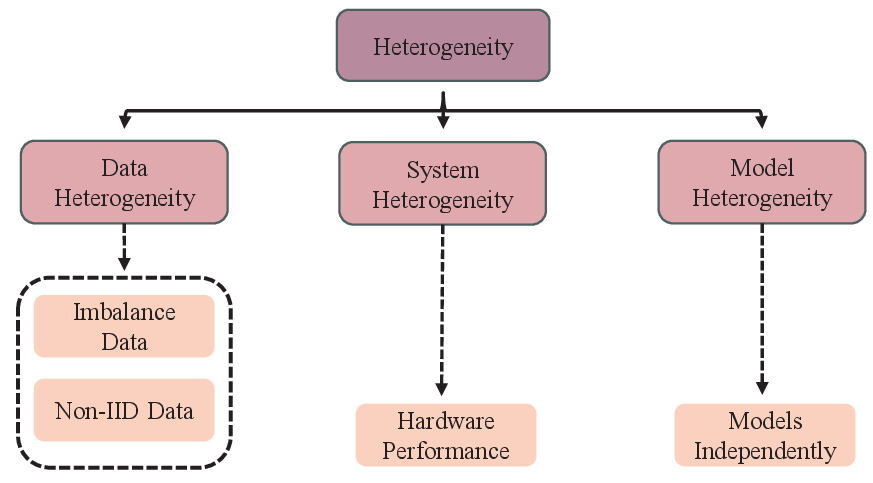}}
  \caption{Main categories of heterogeneity challenges.}
  \label{fig6}
\end{figure}

\subsection{FD for Heterogeneity}
\label{sec4.1}

FL encounters diverse challenges, with heterogeneity emerging as a primary concern due to variations in the learning process. As depicted in Fig.~\ref{fig6}, heterogeneity poses a significant impact on model performance, manifesting in three main categories: 1) \textbf{data heterogeneity}: arises when data across client devices lack independent and identically distributed characteristics; 2) \textbf{system heterogeneity}: contributes to performance constraints on client hardware devices; 3) \textbf{model heterogeneity}: stems from varied client requisites, requiring customized model designs.

While aggregating local models from client devices can alleviate heterogeneity issues in FL \cite{han2022fedx}, effectively managing heterogeneity while preserving model performance remains a challenge \cite{verma2019approaches}. To address this challenge, various methods have been proposed by clustering training data \cite{caldas2018federated,sattler2020clustered},
performing multi-stage training and fine-tuning of models \cite{jiang2019improving},
using brokered learning \cite{fung2018dancing}, distributed multitask learning \cite{corinzia2019variational,smith2017federated} and edge computing \cite{liu2020client}.
Their goals typically involve increasing the volume of training data, mitigating the impact of data skewness, and maintaining a distributed training approach. To tackle system heterogeneity, certain methods focus on balancing tasks and resources among client nodes during training \cite{dinh2020federated,li2019fedmd}. Moreover, addressing model heterogeneity entails assigning accurate weights to client-side models in collaborative training on the client side. Notable methods for this purpose include robust FD \cite{muhammad2022robust}, personalized FD \cite{zhang2021parameterized,hu2021mhat}, and heterogeneous ensemble-based FD \cite{cho2022heterogeneous}.

\subsubsection{Data Heterogeneity}
\label{sec4.1.1}
\

In the context of FL, individual clients possess distinctive sets of training data, often characterized by discernible data distributions \cite{sattler2019robust}. Specifically, the conditional distribution $P(X \mid Y)$ of private data may markedly vary from one client to another, even when a common $P(Y)$ is shared. Consequently, the local data housed at each client does not accurately represent the overall global data distribution. This phenomenon is commonly acknowledged as data heterogeneity, encompassing several key characteristics:

\begin{itemize}[]

  \item {\bf Imbalanced Data}:
        The training data distribution among individual clients exhibits significant variability, primarily arising from diverse usage patterns among users of client devices or applications. This variance leads to differences in the amounts of locally collected training data. Such as for medical data, this heterogeneity is particularly pronounced due to factors like domain shifts, pathology imbalances, multimodal data, annotation noise, and varied client tasks \cite{yeganeh2020inverse}. To tackle this challenges, various approaches have been employed, including regularizing knowledge transfer \cite{wang2023novel}, synthetic auxiliary data \cite{huang2023federated,yin2023syncpfl} and self-distillation calibration \cite{li2023federated}.

  \item {\bf Non-IID Data}:
         The training data for a specific client is typically derived from the usage of a mobile device by a particular user. However, in FL, the distribution of training data across clients can vary significantly, and data from one client may not accurately represent the data distribution of other clients \cite{zhao2018federated, kairouz2021advances}. To address this challenge, various approaches can be employed. One option is to augment the data, making it more similar between clients. This could involve creating a small dataset that can be shared globally, sourced from a publicly available proxy data source or a non-sensitive dataset of client data, potentially distilled from the original data \cite{wang2018dataset}. Another strategy involves building on existing data for each client through collaborative distillation \cite{ahmad2022fedcd}. This approach transforms the Non-IID  problem into a data feature, using client identity to parameterize the model and mitigate some of the challenges posed by Non-IID distributions.

\end{itemize}

While FL indirectly accesses device data, the inherent heterogeneity of data across devices introduces challenges in the convergence of the global model. To address this issue, recent FD methods have also explored different strategies to optimize the local model.
Given that current algorithms are primarily designed to address the heterogeneity caused by Non-IID data, we focus on relevant FD algorithms specifically designed for this aspect as follows.\\

\begin{figure}[!ht]
\centering
\includegraphics[width=3in]{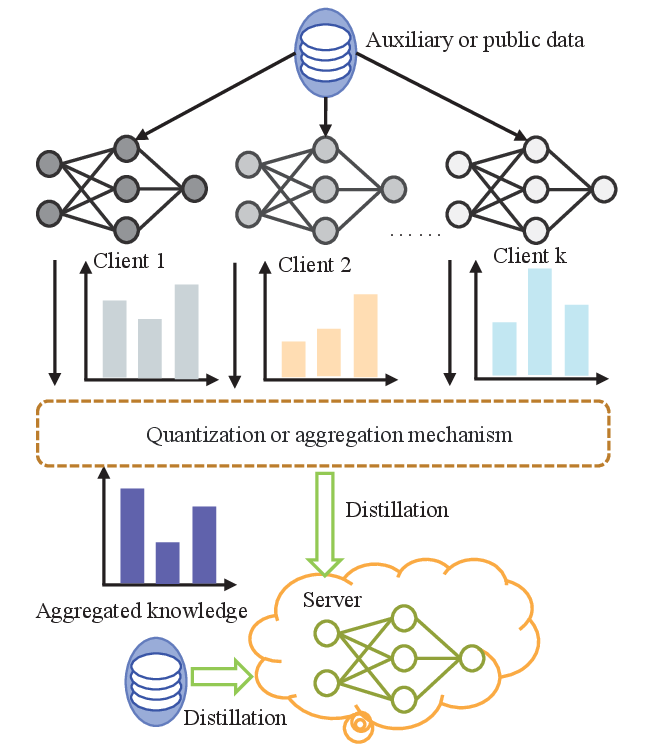}
\caption{The main FD framework with public data.}
\label{fig7}
\end{figure}

\noindent\textbf{FD with Public Data}. By using a public or auxiliary dataset, it incorporates extra information to handle data differences across clients, thus facilitating the global model for better convergence in the FD framework \cite{witt2021reward,nguyen2022cdkt,li2022incentive,chen2023confidence,he2023dealing,li2023fedh2l}. The architecture showing how the auxiliary dataset is involved in training is depicted in Fig.~\ref{fig7}. Specifically, DS-FL transfers local model knowledge by the unlabeled public dataset and proposes entropy reduction averaging methods to aggregate logits information at the server \cite{itahara2021distillation}. FedCM \cite{huang2021federated} updates client-side knowledge by mutual distillation, in which the client uploads predicted logits on public data and cross-entropy information on private test data to the server, the clients receive from the server the set of logits and cross-entropy information without their own information. While Li et al. \cite{li2023incentive} proposes to introduce a public dataset during the local training process, where local models learn knowledge from other local models through a collaborative distillation mechanism. Selective-FD \cite{shao2023selective} filters misleading and ambiguous knowledge in the uploaded information by screening anomalous samples based on an auxiliary dataset, then return the server aggregation information to the client for KD. Another method FEDIC uses auxiliary data to fine-tune the global model to mitigate the bias caused by client data uncertainty, and utilizes ensemble distillation to further optimize the global model by extracting unbiased knowledge from the local model logits optimized with correction mechanisms \cite{shang2022fedic}. FedBiKD proposes a bidirectional distillation mechanism that applies unlabeled public dataset to the server and extracts the knowledge of the global model to guide the local training to alleviate the local bias problem. Simultaneously, aggregating local model knowledge fine-tunes the global model to reduce fluctuations in training \cite{shang2023fedbikd}. These strategy enhances the model's robustness in the face of heterogeneous data distributions. However, when the distribution of the public dataset diverges from the global distribution, relying on the knowledge extracted from the public dataset for guiding model training might lead to a reduction in the overall accuracy of the global model.\\

\begin{figure}[!t]
  \centering
  {\includegraphics[width=3.5in]{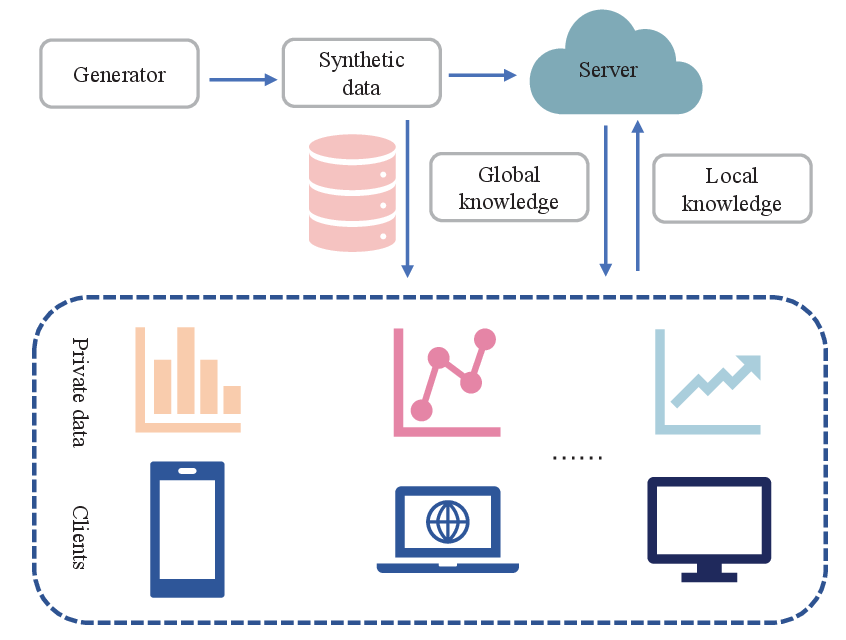}}
  \caption{The main FD framework with synthetic data, which is also known as data-free FD.
  }
  \label{fig8}
\end{figure}

\noindent \textbf{FD with Synthetic Data}. Since a public dataset matching the distribution of a specific application context might not always be available, various approaches have explored different methods to create a auxiliary dataset using private data sources \cite{zheng2023data,peng2023fedgm}. The main FD framework with synthetic data is shown in Fig.~\ref{fig8}, which is also known as data-free FD. Specifically,
FedFTG utilizes pseudo data for fine-tuning the global model~\cite{zhang2022fine}, while Fed2KD tackles distribution of Non-IID data across devices by creating a local IID dataset on each device~\cite{wen2023communication}.
FedDA \cite{wen2022communication} tackles Non-IID data in the FD framework by using conditional autoencoders for data augmentation. Meanwhile, FedBKD \cite{qi2022fedbkd} and FHFL \cite{chen2023privacy} create synthetic datasets covering global data categories for model training, reducing data heterogeneity impact by using KD instead of traditional aggregation. FedF$^2$DG's global model fully learns knowledge from pseudo-data generated by adaptive data generation mechanism through KD to tackle the problem of label distribution shift in Non-IID \cite{zhao2023data}. And SCCD \cite{mai2023server} proposes to generate data based on Gaussian distribution, while proposing a model fusion approach for server-client co-distillation to further address data heterogeneity by extracting key knowledge from localized models. For FedSND \cite{zheng2023federated} and FedNKD \cite{zhu2022fednkd} generate auxiliary data by utilizing random noise, and alleviate the instability resulting from non-IID data by extracting high-level client knowledge to the server using the KD mechanism.
These methods obviate the need for data sharing while striving to mitigate the impact of data heterogeneity in the local model, a prerequisite for the global model's efficacy.

Recently, FAug \cite{jeong2018communication},  FedDTG \cite{zhang2022feddtg} and FRL \cite{zhou2023digital} further explores GANs within the server to grasp the underlying data distribution of clients. FedACK \cite{yang2023fedack} introduces an adversarial contrastive-based KD mechanism to extract global data knowledge into individual local models.
These data-free FD methods, which do not require any public or auxiliary dataset, attempt to mitigate the effects of data heterogeneity in the local model, while the performance of the global model remains bounded by the effectiveness of the local model. Meanwhile the generator model that produce synthetic data need to be trained on Non-IID data. As a result, generator models suffer from inherent data heterogeneity. Therefore, follow-up studies should focus on to avoid the potential impact of data heterogeneity itself in a data-free FD framework.\\

\begin{table*}
\centering
  \caption{Summarising existing FD algorithms for Data Heterogeneity. Heterogeneity type is denoted as HT, Public dataset is denoted as PD, Experimental dataset is denoted as ED. }\label{tb1}

  \begin{tabular}{c|c|c|c|c} \hline
  Methods                  &Technology             &HT            & PD          &ED  \\ \hline

  Selective-FD \cite{shao2023selective} &KD & Non-IID       & Need & Mnist/Femnist, Cifar10 \\ \hline
   CDKT-FL \cite{nguyen2022cdkt} & KD  & Non-IID    & Need  &FmnistT/mnist, Cifar10/100\\ \hline
  Li et al. \cite{li2023incentive} &KD & Non-IID       & Need   &  Mnist/Femnist, Cifar10/100\\ \hline
  DS-FL \cite{itahara2021distillation} & KD  & Non-IID       & Need  &Fmnist/Mnist, Imdb, Reuters  \\ \hline
  FedReID \cite{zhuang2020performance} & KD    & Non-IID       & Need    &Msmt17, Dukemtmc-reid, et.al.\\ \hline
  FedBiKD \cite{shang2023fedbikd} & Bidirectional KD & Non-IID       & Need &Sfd3  \\ \hline
  FEDIC \cite{shang2022fedic}    &Ensemble KD   & Non-IID       & Need   &Cifar10/100LT, ImageNetLT\\ \hline
  Li et al. \cite{li2022incentive}    & Incentive mechanism  & Non-IID       & Need  & Femnist/mnist, Cifar10/100 \\ \hline
 PTSFD \cite{witt2021reward}    & 1-bit compressed KD   & Non-IID       & Need   &Emnist/mnist\\ \hline
  He et al. \cite{he2023dealing} &Dual KD & Non-IID       & Need &Knee MRI \\ \hline
   FedH2L \cite{li2023fedh2l}    & Mutual distillation   & Non-IID       & Need    &Minist, Pacs, Office-Home\\ \hline
  FedCM \cite{huang2021federated} &Mutual distillation & Non-IID       & Need  & Adni, Oasis-1, Aibl\\ \hline
  Chen et al. \cite{chen2023confidence}  &Confidence-based KD & Non-IID       & Need  &Udacity \\  \hline

  FedCRAC \cite{li2023federated}  & Self-distillation  &Imbalance  &None & Iscx, Quic  \\ \hline
  Fedkt \cite{wang2023novel} & KD & Imbalance          &None  & Hmeq, Gmsc, et al. \\ \hline
  SynCPFL \cite{yin2023syncpfl} & KD & Imbalance          &None  &Fmnist/Mnist, Cifar10 \\ \hline
  FedLGD \cite{huang2023federated}&Local-Global distillation  & Imbalance          &None  & Mnist, Svhn, et.al.\\ \hline

  FedBIP \cite{zhang2023fedbip}  &  KD   &Non-IID     & None  &  Wind turbine dataset\\ \hline
  FedBE \cite{chen2020fedbe} &  KD   &Non-IID     & None & Cifar10/100, Tiny-ImageNet  \\  \hline
  FedNKD \cite{zhu2022fednkd} &  KD   &Non-IID     & None & Fmnist, Cifar10, Svhn  \\  \hline
  FedDA \cite{wen2022communication} &  KD   &Non-IID     & None & Fmnist, Cifar10  \\  \hline
  PersFL \cite{divi2021unifying} &  KD   &Non-IID     & None & Cifar10, Mnist   \\  \hline
  FedNoRo \cite{wu2023fednoro} &  KD   &Non-IID     & None & Ich, Isic2019   \\  \hline
  FedBR \cite{zeng2023enhanced} &  KD   &Non-IID     & None & Cifar10/100   \\  \hline
  AFSKD \cite{feng2023adapter}  &  KD   &Non-IID     & None & Qmsum  \\  \hline
  UNIDEAL \cite{yang2023unideal}  &  KD   &Non-IID     & None & Mnist/Mnist-M, et al.   \\  \hline
  GFedKRL \cite{ning2023gfedkrl} &  KD   &Non-IID     & None & Mutag, Bzr, et al.   \\  \hline
  FHFL \cite{chen2023privacy} &  KD   &Non-IID     & None &  Mnist, Lung Xary/CT   \\  \hline
  LSG \cite{bai2023overcoming}  &  KD   &Non-IID     & None &Fmnist/Mnist, et al.\\   \hline
  FedHKD \cite{chen2023best}   &KD     &Non-IID      & None  & Svhn, Cifar10/100  \\ \hline

  FedFTG \cite{zhang2022fine} &Data-free KD     &Non-IID      & None  & Cifar10/100\\ \hline
  FedSND \cite{zheng2023federated} &Data-free KD   &Non-IID   & None  &Fmnist, Cifar10, et al. \\ \hline
  FedDTG \cite{zhang2022feddtg} &Data-free KD     &Non-IID      & None &Fmnist/Emnist/Mnist \\ \hline
  Zheng et al. \cite{zheng2023data}  &Data-free KD     &Non-IID      & None    & Rnsa-baa  \\ \hline
  FedF$^2$DG \cite{zhao2023data} &Data-free KD     &Non-IID      & None &Svhn, Cifar10/100  \\ \hline

  $CD^2$-pFed \cite{shen2022cd2} & Cyclic KD    &Non-IID      & None & Cifar10/100, et al. \\ \hline
  DaFKD \cite{wang2023dafkd}  &Domain-aware KD  &Non-IID      & None  &Fmnist/Emnist/Mnist, Svhn \\ \hline
  FRL \cite{zhou2023digital} &Bidirectional distillation  &Non-IID      & None &Mnist, Cifar10\\ \hline
  FedX \cite{han2022fedx}   & Two-sided KD   &Non-IID      & None  &Fmnist, Cifar10, Svhn\\ \hline
  Fed2KD \cite{wen2023communication}  &Two-step KD  &Non-IID      & None   & Fmnist, Cifar10, ImageNet\\ \hline
  FedACK \cite{yang2023fedack}  & Adversarial contrastive KD   &Non-IID      & None &  Vendor-19, TwiBot-20  \\ \hline
  FedLN \cite{tsouvalas2022federated}    & Adaptive KD   &Non-IID      & None &Fmnist, Pathmnist, et al.  \\ \hline
  FedDAT \cite{chen2023feddat}    & Mutual distillation    &Non-IID      & None  & VizWiz, Art, et al.\\ \hline
  FedCSR \cite{zhang2023cuing}   & Mutual distillation    &Non-IID      & None  &Chinese CS dataset\\ \hline
  FedGM \cite{peng2023fedgm}     & Mutual distillation   &Non-IID      & None  &Emnist/Mnist, Cifar10\\ \hline
  Fed2Codl \cite{ni2022federated}     &Mutual distillation  &Non-IID      & None &  Femnist, Cifar10, et al.  \\ \hline
  EFDLS \cite{xing2022efficient}   &Weights matching, FD  &Non-IID      & None  &Ucr2018 \\ \hline
  LSR \cite{jiang2022towards} & Self-distillation  &Non-IID      & None & Fmnist/Mnist, Cifar10\\ \hline
  FedBKD \cite{qi2022fedbkd}    & Bidirectional KD    &Non-IID      & None   &Qpsk, 4Fsk, et.al\\ \hline
  SCCD \cite{mai2023server}    & Collaborative distillation   &Non-IID      & None  &CartPole, Pendulum, et.al\\ \hline
  FedDKD \cite{li2023feddkd}    & Decentralized KD   &Non-IID      & None  &Femnist/Emnist, CIFAR10/100\\  \hline
  MrTF \cite{li2023mrtf} &Rectified distillation   &Non-IID      & None  & Mnist/Mnistm, et al.\\ \hline
  FAug \cite{jeong2018communication} & Online KD  &Non-IID      & None & Mnist\\ \hline
  ConDistFL \cite{wang2023condistfl}  & Conditional distillation  &Non-IID      & None & Msd, Kits19, Amos22\\ \hline
  \end{tabular}
\end{table*}

\noindent\textbf{FD with Global Alignment}. Besides public or synthetic data, several methods have recently been introduced  to make the local knowledge consistent with the global knowledge \cite{ni2022federated,tsouvalas2022federated,wang2023condistfl}. One strategy is to rely on global knowledge constraints \cite{ning2023gfedkrl,wu2023fednoro}. PersFL introduces knowledge transfer by selecting optimal teacher models during the FL training phase, benefiting each local model \cite{divi2021unifying}. MrTF uses a global initial model and a global model aggregated in each training round to jointly constrain local models \cite{li2023mrtf}. FedHKD adjusts local models training through the integration of server-aggregated hyper-knowledge, comprising soft predictions and data representations corresponding to each other \cite{chen2023best}. Additionally, ASFKD proposes a global knowledge-based entropy selection strategy adapting to constrain local models by extracting knowledge from a global adapter \cite{feng2023adapter}. FedBR designs block-wise regularization, which absorbs knowledge from global model blocks via KD \cite{zeng2023enhanced}.

Another strategy like UNIDEAL introduces adjustable mutual evaluation curriculum learning mechanism to constrain the local model training by a global model \cite{yang2023unideal}. Similarly, Fed2Codl \cite{ni2022federated}, $CD^2$-pFed \cite{shen2022cd2}, FedDAT \cite{chen2023feddat} and FedCSR \cite{zhang2023cuing} utilizes a mutual distillation mechanism to prevent local models from biased training on non-IID data.
These studies underscore that the server possesses ample information, and the aggregated global model demonstrates competitive performance, preventing clients from converging to locally optimal models due to Non-IID circumstances.\\

\noindent\textbf{FD with Local Alignment}. To mitigate the influences of imperfect data on local model training, recent methods also explored the knowledge alignment from the perspective of local clients, including self-distillation, and inter-client distillation. Specifically, LSR \cite{jiang2022towards} and LSG \cite{bai2023overcoming} propose local regularization and local self-guiding strategies to train local models to upload important client information to the server, respectively.
Similarly, EFDLS proposes to transfer inter-client hidden layer features with feature-based KD and proposes a weight matching scheme to update hidden layer weights \cite{xing2022efficient}. FedBIP proposes optimizing the local model using KD and constraining the global model update based on the update direction of the majority of the participating trained local models \cite{zhang2023fedbip}.
Moreover, DaFKD \cite{wang2023dafkd} introduces a domain-aware mechanism to enhance the importance of specific distillation samples within each local model. FedDKD extracts the local model neural network graph mean knowledge by KD for global model training, which effectively copes with the damage caused by local data heterogeneity \cite{li2023feddkd}. These methods mitigate data heterogeneity by optimizing the uploading of local model information.\\

\noindent\textbf{FD with Hybrid Strategies}. To get better model performance, multiple strategies are engaged in model training to cope with data heterogeneity. For instance, FedBE proposes to use Bayesian inference to sample high-quality local models for more robust aggregation, and the global model constrains the local model with unlabeled data predictions via KD \cite{chen2020fedbe}. FedReID fine-tuned the server model using client-generated knowledge on public datasets and dynamically adjusted the backbone network of the local model using the global model \cite{zhuang2020performance}. FedX introduces a dual KD mechanism, which employing local KD training the network progressively based on local data and global KD regulating local model to effectively address data bias in Non-IID scenarios \cite{han2022fedx}.  \\

\noindent\textbf{Summary}. Different FD algorithms have emerged to cope with data heterogeneity, but they can be essentially categorized into two types: (1) Using public datasets, generated auxiliary data and global knowledge alignment to enable local models to obtain global knowledge. However, the overall performance of the model will be degraded if the public dataset or the generated dataset differs significantly from the true global data distribution. The global knowledge alignment strategy is also constrained by the uploaded information of the local model, and the unified model is obviously not applicable to all clients, resulting in its limitation in coping with data heterogeneity. (2) Using local knowledge correction aims at extracting locally important information, but data information with small amount of data is easily ignored, resulting in a lack of generality of the global model. The existing algorithms lack the exploration of data heterogeneity in which cross-domain dataset or multi-task dataset appear. And the imbalance of class samples or even the absence of special class samples in client-side heterogeneous data is also worth considering, which exists in real-world environments, such as the different distribution of records of different diseases in hospital. Hence, for enhancing the robustness of the FD framework, mitigating data heterogeneity from multiple perspectives will be the focus of subsequent research.

\begin{figure}[!ht]
\centering
  \includegraphics[width=3.5in]{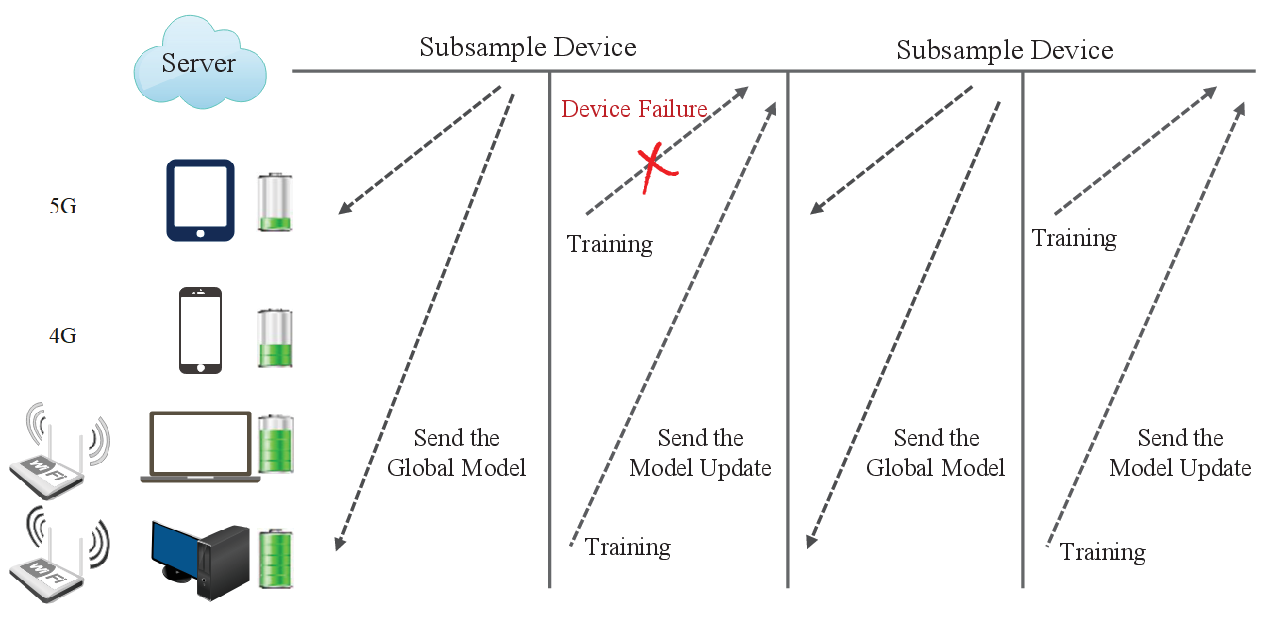}
  \caption{
  Heterogeneity in System Environments for FD Models. Devices exhibit variations in terms of network connectivity, power supply, and hardware performance. Moreover, certain devices might experience dropout during the training process. Hence, the FD framework necessitates the ability to accommodate diverse system environments and address the challenge of low device participation.
  }
  \label{fig9}
\end{figure}

\subsubsection{System Heterogeneity}
\label{sec4.1.2}
\

System heterogeneity describes the variances in the capabilities of client devices, including storage, computation, and bandwidth for communication, as illustrated in Fig.~\ref{fig9}. Due to factors like hardware performance, network connectivity, stability and power supply fluctuations, some clients might not complete their local training within the specified time frame, potentially affecting overall aggregation performance. The existence of system heterogeneity—where participants possess varying bandwidth and computing power—has gained substantial attention from the community~\cite{li2019fedmd,li2020federated}.

A simple and straightforward way to address system heterogeneity involves directly removing weak clients without sufficient resources during training~\cite{bonawitz2019towards}. However, this will also increase the risk of missing important data on weak clients, resulting in the loss of global model performance. Alternatively, a small global model can be designed based on the weakest client as a baseline to satisfy all clients' participation in training, but the performance of the global model is easily limited by its architecture, which makes it difficult to be applied in reality. In contrast, it is easy to cope with above problem in the FD framework. To address the above mentioned system heterogeneity, recent FD solutions are grounded in two key considerations: (1) adjusting the model in response to the performance of client devices; and (2) maximizing the utilization of available resources. We delve into the specific methodologies as follows.\\

\noindent\textbf{FD with Model Adjustment}. The first approach is to compress the local model using KD based on the client's performance. Clients adopt lightweight models to mitigate the constraints imposed by device hardware. NRFL \cite{mishra2021network} utilizes the KD compression local model to adapt to different bandwidth requirements. Jain et al. \cite{jain2021federated} introduces a compression model of the intermediate assistant teacher-based KD mechanism to alleviate the computational resource constraints. Tanghatari et al. \cite{tanghatari2023federated} proposes to reduce the computational requirements by utilizing heuristic method to introduce the FD framework to generate small models by trimming the original model to participate in training. FedGKT addresses the limitations of edge computing by employing an alternating minimization strategy to train lightweight networks on the edge \cite{he2020group}. ScaleFL \cite{ilhan2023scalefl} suggests an adaptive downscaling of the global model along both width and depth dimensions based on the computational resources available from participating clients. The local model aggregation updates, originating from heterogenous clients, are achieved via an improved self-distillation process, which enhances the knowledge transfer across subnetworks. These methods mainly consider alleviating the constraints of a particular aspect of the client device and lack consideration of different constraint situations. Meanwhile, they create a huge risk to the privacy and security of the client's private information due to the shared partial modeling structure.\\

\noindent\textbf{FD with Maximum Available Resources}. Another approach that is more fully considered is to design the model based on the client's resources \cite{ozkara2021quped,yuan2023hetefedrec,shi2022closing,hamood2023clustered}. For instance, In order to fully utilize the capabilities of all devices and their hardware, a novel training architecture called FjORD has been introduced. FjORD mitigates the negative impacts of client system heterogeneity by integrating an ordered dropout mechanism and a self-distillation strategy, which dynamically adjusts the model size according to the client's capabilities \cite{horvath2021fjord}. And InclusiveFL assigns models based on the computational power of the clients, with powerful clients assigned large models and weaker clients assigned small models. Knowledge from the large models of powerful clients is transferred to the small models of weak clients using the momentum KD technique \cite{liu2022no}. Similarly, FedZKT designs local models independently based on the local resources of the client device from which they are derived, and achieves knowledge transfer of models across heterogeneous devices through a zero-sample distillation approach \cite{zhang2022fedzkt}. On the other hand, Wang et al. \cite{wang2023digital} proposes a digital twin based FD framework to cope with the limitation of client resources by Q-learning to select different models for the clients and utilizing KD to train a copy of the client model at the server. Chen et al. \cite{chen2023resource} proposes to group clients based on their computational power and communication constraints, and design models based on client performance to ensure that every client participates in training. However, consider that some clients may not be involved in the entire model training process, resulting in loss of critical information. FedMDP proposes using KD and dynamic local task allocation mechanism to ensure that a client-based design model does not quit during training due to its own resource constraints \cite{imteaj2023fedmdp}.\\

\noindent\textbf{Summary}. The above FD algorithms have achieved certain achievements in coping with the system heterogeneity problem. However, when the FD framework is extended in real-world scenarios, along with the increasing complexity of the client, there will be multiple constraints constraining the training of the model at the same time. For example, different clients have large differences in the sizes of the private data they can accommodate, while the communication bandwidth and communication time requirements are different during model training. In this case, the limitation of local model size based on different client resources by compressing and fixing the model design appears. When designing local models based on client resources, how to balance multiple client constraints to select a suitable network will also be a new challenge.

\begin{table*}[!t]
\centering
  \caption{Comparison of FD approaches to address Model Heterogeneity, wherein knowledge distill represents information communication between models by KD extracting model logits and the default corresponding weight is one, while optimize aggregated information denotes the optimization weight of uploaded logits information for the local model.}\label{tb2}

\begin{tabular}{c|c|c|c} \hline
  Methods & Role of KD in FL  & Application domain  & Dominant strategy  \\ \hline

  FedGEMS \cite{cheng2021fedgems}   & Optimize aggregated information &Image classification  &KD  \\\hline
  DFRD \cite{luo2023dfrd} &Optimize aggregated information  & Image classification & KD \\  \hline
  KT-pFL \cite{zhang2021parameterized}  &Optimize aggregated information  & Image classification  &  KD \\ \hline

  HBS \cite{wang2020industrial} & Optimize aggregated information   & Image classification  & Bidirectional KD \\ \hline
  FEDAUX \cite{sattler2021fedaux}    & Optimize aggregated information   & Image classification         &Ensemble distillation\\\hline
  COMET \cite{cho2023communication} &Optimize aggregated information  &Image classification   & Clustered co-distillation \\     \hline
  FedCD \cite{ahmad2022fedcd}   &Optimize aggregated information  &Image classification   & Collaborative KD \\ \hline

  Gudur et al. \cite{gudur2020federated}  & Optimize aggregated information   &  Activity recognition  & Bidirectional KD \\ \hline
  RMMFL \cite{muhammad2022robust} & Optimize aggregated information & \begin{tabular}{c}Digit classification, \\ Image  classification \end{tabular}  & KD \\ \hline
  FedKEM \cite{nguyen2023enhancing} & Optimize aggregated information &\begin{tabular}{c}Digit classification, \\ Image classification \end{tabular}   & KD \\ \hline
  Fed-ET \cite{cho2022heterogeneous} & Optimize aggregated information & Sentiment classification  & Ensemble KD \\ \hline
  SQMD \cite{ye2023heterogeneous} &Knowledge distill  & Healthcare Analytics  &  Messenger KD \\   \hline
  Cronus \cite{chang2019cronus} &Knowledge distill  & \begin{tabular}{c}Digit classification, \\ Image classification \end{tabular}   & KD \\  \hline
  FedZKT \cite{zhang2022fedzkt} & Knowledge distill &\begin{tabular}{c}Digit classification, \\ Image classification \end{tabular}   & Zero-shot KD \\\hline
  FedDF \cite{lin2020ensemble}  & Knowledge distill &\begin{tabular}{c}Image classification, \\ News classification \end{tabular}  &Ensemble KD \\\hline
  FHFL \cite{chen2023privacy}   & Knowledge distill & \begin{tabular}{c}Medical classification, \\ Image  classification \end{tabular}  & KD  \\ \hline
  FedBKD \cite{qi2022fedbkd} & Knowledge distill & Signal classification  & Bidirectional KD \\ \hline
  FML \cite{shen2023federated} &Knowledge distill & Image classification  & Mutual learning \\ \hline
  MHAT \cite{hu2021mhat} & Knowledge distill & Image classification  &KD  \\ \hline
  FedMD \cite{li2019fedmd} &Knowledge distill  &Image classification   &KD  \\ \hline
  FAS \cite{liu2022communication} & Knowledge distill & Image classification  & KD \\ \hline
  FedGD \cite{zhang2023towards}   & Knowledge distill & Image classification  & KD  \\ \hline
  DENSE \cite{zhang2022dense} &Knowledge distill  & Image classification & KD \\  \hline
  FedHE \cite{chan2021fedhe} &Knowledge distill &Image classification & KD\\ \hline

\end{tabular}
\end{table*}

\subsubsection{Model Heterogeneity}
\label{sec4.1.3}
\

Model heterogeneity refers to model inhomogeneity, which usually means different model sizes or structures.
The majority of existing studies in FL, such as FedAvg \cite{mcmahan2017communication}, employ a uniform model training approach for all clients. However, the unique requirements of local models in real-world scenarios often differ across clients, leading to sub-optimal performance of a single global model across all clients. For example, several healthcare organizations collaborate without sharing private data, and they may need to develop their own models to meet different specifications. Participants independently tailor their models for different demands, resulting in different model sizes or inconsistent backbone model choices (e.g., ResNet \cite{he2016deep}, EfficientNet \cite{tan2019efficientnet}, and MobileNet \cite{howard2017mobilenets}). Therefore, constructing personalized local models based on client hardware performance constraints and requirements exposes the FL framework with model heterogeneity challenges. Local models with different model architectures cannot be directly aggregated or summarized in the server. It is therefore not feasible to transfer model gradients or parameters between the client and the server for model training.

To address this challenge, an emerging and promising solution that has garnered substantial attention involves enabling devices to autonomously design their own on-device models within a FD framework. This approach allows clients to break free from the constraints of standardized model architecture, effectively addressing the constraint of model heterogeneity while still ensuring satisfactory model convergence accuracy. In this approach, local models transfers information that is not restricted by the model structure (e.g., logits) under KD methodology.
A summary of FD techniques designed to tackle model heterogeneity is presented in Table~\ref{tb2}. There follows the summary mainly based on the manner in which the local model logits information was handled.\\

\begin{figure}[!ht]
\centering
\subfloat[]{\includegraphics[width=3in]{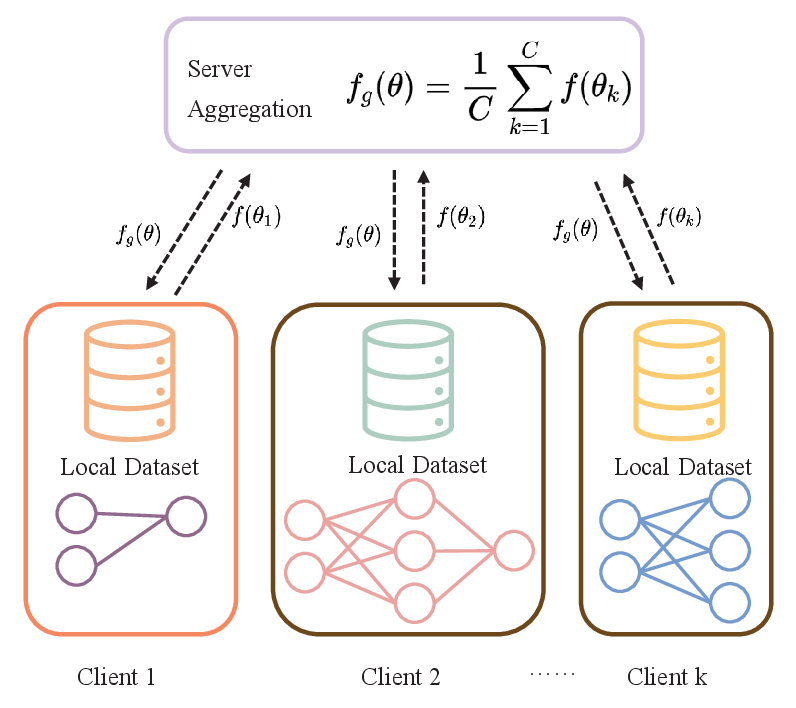} \label{fig10.1}}
\hfil
\subfloat[]{\includegraphics[width=3in]{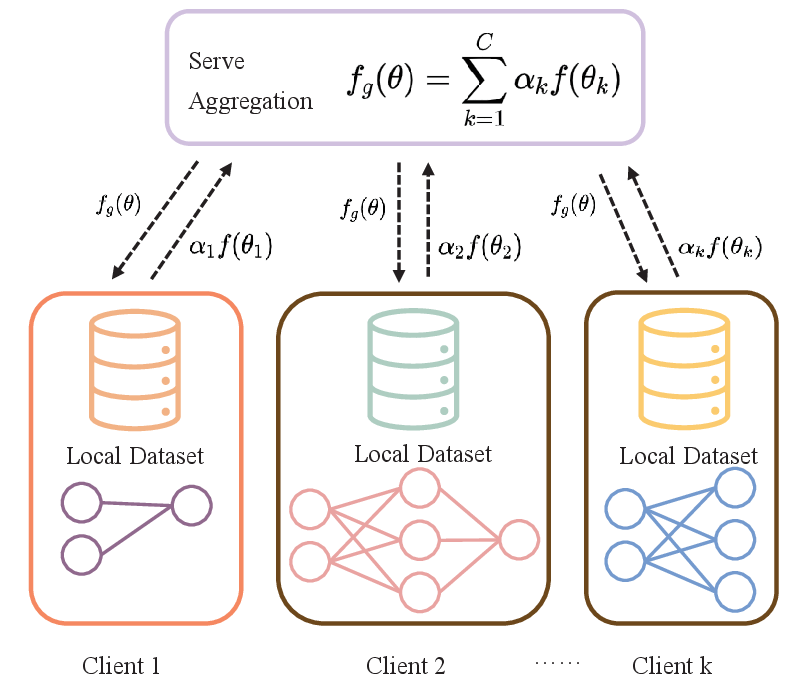} \label{fig10.2}}
\caption{The way servers are aggregated in FD, where (a) shows the server performs straightforward mean-aggregation operation on the client's uploaded model information and (b) is a weighted aggregation based on the contribution of local model information through the weighting mechanism introduced by the corresponding framework.}
\label{fig10}
\end{figure}

\noindent\textbf{FD with Direct Aggregation.} The direct average aggregation of the uploaded logits information from local models at the server is a simple and effective strategy \cite{chang2019cronus}. A visual representation of the frame structure plan is illustrated in Fig.~\ref{fig10.1}. 
The formulation of this direct aggregation approach can be expressed as follows:
\begin{equation}\label{eq9}
        f_{g}\left ( \theta   \right )   =\frac{1}{C}\sum_{k=1}^{C} f\left ( \theta_{k}  \right ) ~,
    \end{equation}
where $f_{g}\left ( \theta \right )$ denotes the information aggregated during a communication round of the server.

The sources of logits uploaded about the local model are mainly categorized as relying on public dataset and synthetic dataset. One category is the transmission of logits using public dataset. Specifically, FedMD and FHFL empowers clients to independently craft their own networks using local private data and a shared public dataset. In this scenario, averaging aggregation of logits information for participating local models to training global model \cite{li2019fedmd,chen2023privacy}. Similarly, FedDF uses server unlabeled data to train the global model by transferring local model average logits through ensemble distillation \cite{lin2020ensemble}. FAS uploads a subset logits with low entropy of each client with the public unlabeled dataset \cite{liu2022communication}.

Another class of algorithms transmits average logits information by generating data. For instance, DENSE utilizes the client-side local model ensemble to train the generator, and the local model as a teacher uses the data synthesized by the generator to transfer the average logits information to the global model \cite{zhang2022dense}. FedGD \cite{zhang2023towards} and FedZKT \cite{zhang2022fedzkt} uses GAN training to generate models to obtain synthetic dataset training models to upload local model average logits information. Similarly, FedBKD \cite{qi2022fedbkd} proposes a bidirectional distillation mechanism to cope with model heterogeneity, while using conditional variational autoencoders to generate auxiliary dataset to enhance model robustness. There is also a rare approach local models do not rely on public dataset or generated dataset, local models upload logits information for private data. Like FedHE transfer logits are derived from a randomly selected batch of private data from the client \cite{chan2021fedhe}.\\

\noindent\textbf{FD with Weighted Aggregation.} Consider the differences in performance and contribution of the local models. Recognizing that the information transmitted by each local model might not lie within the same range, it becomes crucial to fully harness the potential of each local model by appropriately weighting the transmitted information \cite{sattler2021fedaux,ahmad2022fedcd}. The structural layout of the weighted aggregation framework is elucidated in Fig.~\ref{fig10.2}, which involves assigning weight coefficients to the aggregated information from each client:
\begin{equation}\label{eq10}
        f_{g}\left ( \theta   \right )   =\sum_{k=1}^{C}\alpha _{k}  f\left ( \theta_{k}  \right ) ~,
    \end{equation}
where $\alpha _{k}$ denotes the weight assigned to the model transfer knowledge of the client $k$.

Specifically, weighted logits using similarity. KT-pFL defines the knowledge coefficient matrix to update the client's prediction results that reinforce the contribution of clients with similar data distribution \cite{zhang2021parameterized}. COMET proposes collaborative distillation to enhance the logits of local model uploads while weighting them according to the similarity of client data distribution \cite{cho2023communication}. For HBS proposes a bidirectional online distillation mechanism to train the model by transmitting logits knowledge of the model, while dynamically changing the aggregation weights of logits using the similarity of gradient features \cite{wang2020industrial}. Additionally, SQMD evaluates the local model quality by logits information and quantifies the similarity between any two clients by computing the local model upload information at the server, and selects high-quality k nearest-neighbor logits for each client to improve its local model training \cite{ye2023heterogeneous}.

Another strategy utilizing model predictions to perform weighting.
Specifically, Fed-ET weights the local model logits according to their variance \cite{cho2022heterogeneous}. Similarly,  FedKEM's server aggregation policy is to select the maximum logits information in all local auxiliary networks \cite{nguyen2023enhancing}. Furthermore, FedGEMS discards unreliable local models, and the information entropy of reliable local model prediction results is used as the weight for uploaded logits \cite{cheng2021fedgems}.
On the other hand, Gudur et al. \cite{gudur2020federated} utilizes the prediction accuracy of the local model on the public dataset as a weight for the local model to upload server information.

Besides the above weighting strategies, there are different weighting methods to optimize the upload information weights. For instance, to enhance the stability of the collaborative learning process and diminish the performance disparities among local models, the Robust Multi-Model Federated Learning (RMMFL) approach has been introduced, drawing from the foundations of the FedMD framework. RMMFL introduces a high entropy aggregation technique to soften the output prediction and finds the optimal weights for uploading logits information through grid search \cite{muhammad2022robust}. And DFRD \cite{luo2023dfrd} weights the local model upload information using the proportion of data involved in client-side local updates.

\noindent\textbf{FD with Auxiliary Networks.} In addition to the above methods, relying on auxiliary networks can also effectively mitigate model heterogeneity. More specifically, FML proposes to design a meme model containing specific parts of the global model to send to clients, the meme model and the local model are mutual learning by logits using private data, and the server averagely aggregates the meme models from all the clients to be used as the global model for the next training \cite{shen2023federated}. And MHAT argues that weighted averaging of uploaded clients' information is not optimal and proposes to train auxiliary models at the server for aggregating uploaded model predictions \cite{hu2021mhat}.

\noindent\textbf{Summary}. From the above review, the FD framework is found to be robust in the face of model heterogeneity. However, currently the related algorithms are mainly used for classification tasks by transmitting logits information. Therefore, the application scope of FD framework is limited. The application of FD framework in different scenarios (eg. speech recognition, image tracking, and anomaly detection.) will be promoted by extracting appropriate information via distillation mechanism without leaking private data, which will be a hot spot for subsequent exploration.

\begin{table}[!ht]
\centering
  \caption{Comparison of KD-based FL approaches to address client-drift, which local information (L-inform.) indicates the involvement of client knowledge in regulating local model updates, while the global information (G-inform.) stand for the use of global model knowledge to correct the local model.}
  \label{tb3}
\begin{tabular}{c|c|c|c} \hline
  Methods            & Public data & L-inform. & G-inform.\\ \hline

  FedMMD \cite{yang2022fedmmd}              & \checkmark       & \checkmark                  &  \ding{55}                \\ \hline
  FedGKD \cite{yao2023fedgkd}                &   \ding{55}       &          \ding{55}                  &   \checkmark             \\ \hline
  FedCAD \cite{he2022class}                 &   \checkmark       &          \ding{55}                  &   \checkmark             \\ \hline
  FedKF \cite{zhou2023handling}             & \ding{55}         &          \ding{55}                  &   \checkmark               \\ \hline
  CADIS \cite{nguyen2023cadis}              &   \ding{55}       &         \ding{55}                 &  \checkmark                   \\ \hline
  FedMLB \cite{kim2022multi}                &   \ding{55}       &        \ding{55}                    &   \checkmark                \\ \hline
  FedAlign \cite{mendieta2022local}        &   \ding{55}       &          \ding{55}                &   \checkmark                  \\ \hline
  FedICT \cite{wu2023fedict}                &   \ding{55}       &         \ding{55}                  &   \checkmark                \\ \hline
  FedLU \cite{zhu2023heterogeneous}         &   \ding{55}       &         \checkmark                  &   \checkmark                 \\ \hline
  FSAR \cite{guo2023fsar}       &   \ding{55}     & \ding{55}          &   \checkmark       \\ \hline
  Yashwanth et al. \cite{yashwanth2023federated}      &   \ding{55}     &    \ding{55}        &  \checkmark      \\ \hline
  KDAFL \cite{shen2023discrepancy}  &   \ding{55}     &   \ding{55}   &   \checkmark    \\ \hline
  FedCSD \cite{yan2023rethinking}      &   \ding{55}     &   \ding{55}   &   \checkmark    \\ \hline
  FEDGEN \cite{zhu2021data} &   \ding{55}     &    \checkmark    &   \ding{55}   \\ \hline
  \end{tabular}
\end{table}

\subsection{FD for Client-drift}
\label{sec4.2}

When clients are trained using their local heterogeneous dataset, the models trained on these local datasets can exhibit a discrepancy between the average outcome of the local optima and the minimum global empirical loss of the global model. This discrepancy indicates that the average model might differ significantly from the global optimum, a phenomenon referred to as “client drift” \cite{kim2022communication}. For example, in the context of collaborative training models for disease diagnosis across multiple hospitals, each hospital possesses its own private patient data. It's possible that while one hospital may have a substantial data sample for a specific disease, others may lack data samples for the same disease. As each hospital performs local training on its own local model, its local objectives could deviate substantially from the global objectives. This implies that during local training, each client optimizes primarily for its local optimal solution, rather than addressing the global objective. Consequently, this phenomenon introduces a drift in the collective client updates, ultimately yielding sub-optimal model performance in the global model.

In addition to the solutions discussed in the previous section, which address heterogeneity issues in data, system, and model, various studies have specifically focused on addressing the challenge of client drift during local training. The literature on this matter is summarized in Table~\ref{tb3}, highlighting different key aspects that researchers have investigated. The root cause of client drift is that the client consistently starts with the same global model, while its local model, trained on private data, gradually moves away from the optimal value of its global objective. Consequently, the aggregated global model does not reach the optimal value of the global objective. That is, the knowledge the client incorporates when training the model is biased. Therefore, we categorize existing methods for addressing the client drift problem into two main aspects.\\

\noindent\textbf{FD with Global Regularization}. One line of work aims to correct imbalances in client information using global model information \cite{yao2023fedgkd}. Specifically,
FedCSD considers the similarity of the local and global information, which aligns the local logits with the global logits by class prototype similarity distillation \cite{yan2023rethinking}. In a similar vein, Yashwanth et al. \cite{yashwanth2023federated} proposes to use the similarity between the global and local model outputs to adjust the weights of the local model trained on private data.
CADIS proposes within a single training cycle, the local model acts as a student learning knowledge from a global model that aggregates rich knowledge to overcome the local optimum problem during training \cite{nguyen2023cadis}. And KDAFL proposes a selective knowledge extraction strategy based on the overall knowledge difference between local and global models, which utilizes the knowledge of the global model to guide the training of local models with poorer learning outcomes \cite{shen2023discrepancy}.

\begin{figure}[!t]
\centering
 \includegraphics[width=3.5in]{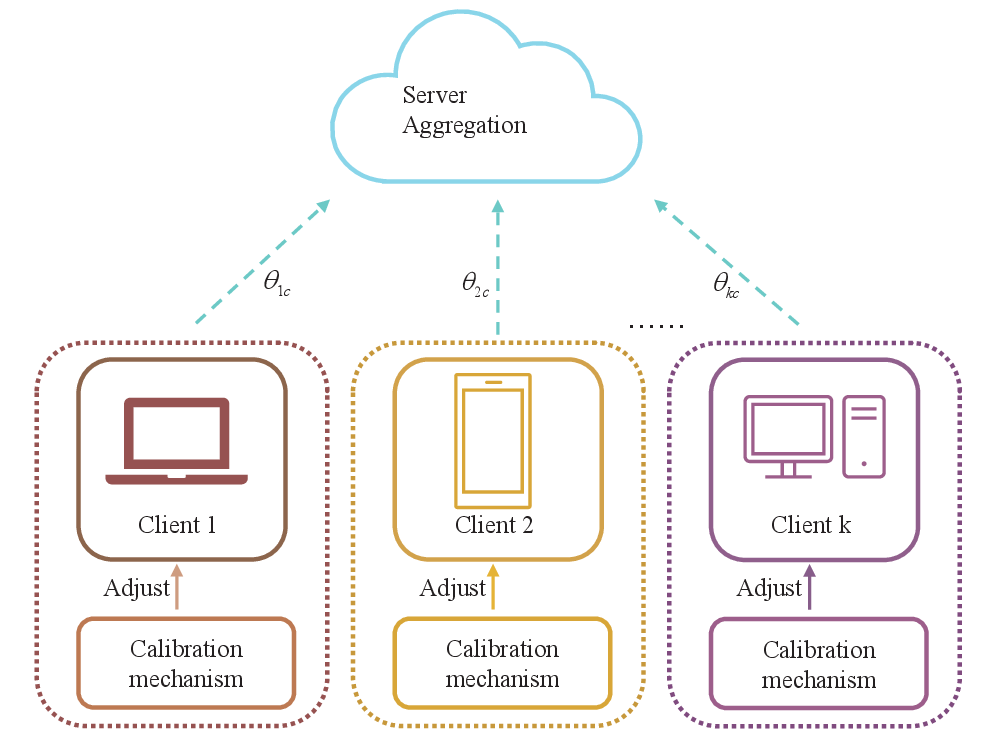}
 \caption{Calibration mechanisms constrain local model training, where $\theta_{1c}$, $\theta_{2c}$ and $\theta_{kc}$ are the local model information passed from the client to the server.}
\label{fig11}
\end{figure}

Similarly, FSAR \cite{guo2023fsar} introduces the Multi-grain KD mechanism, which employs the shallow features of the global model as teacher knowledge to guide the local training of individual clients. FedMLB utilizes the global subnetwork module in conjunction with the local network to construct multiple auxiliary branches and learns the knowledge of each branch through online distillation to reduce the deviation of the feature space of the local model from the global model \cite{kim2022multi}. And FedKF \cite{zhou2023handling} generates pseudo-samples by fusing client-side knowledge through the global-local knowledge fusion technique, employing KD to have the student network learn from the teacher network. In each training round, the server imparts knowledge fused with local knowledge to guide the local model's training towards the global optimum. Incorporating a calibration mechanism enhances the utilization of global model representation knowledge, facilitating local model convergence and its alignment with the global model throughout the training process. This global calibration mechanism is integrated into the FD framework to fine-tune the local model training process, as illustrated in Fig.~\ref{fig11}. The general formulation of the calibration function for a local model can be expressed as follows:
\begin{equation}\label{eq11}
L_{k}^{reg}  =\frac{1}{B}\sum_{i\in \left [ B \right ]   } \alpha _{ki}\left ( x_{kj},y_{kj}  \right )   L_{KL} \left ( q_{g}\left ( x_{kj} \right ), q_{k}\left ( x_{kj} \right )  \right )~,
\end{equation}
where $B$ is denoted the batch size, $\alpha _{ki}\left ( x_{kj},y_{kj}   \right )$ is the weight for the sample $x_{kj}$ with ground truth label $y_{kj}$, and $q_{g}\left ( x_{kj} \right )$ and $q_{k}\left ( x_{kj} \right ) $ are logits on the temperature scale of the global model and local model $k$ respectively.\\

\noindent\textbf{FD with Adaptive Regularization}. Using the global model to force corrections suppresses drift but also limits the potential for local convergence. In this way, not as much new information is collected in each round of communication. FedAlign proposes to constrain key local network modules to promote local models to learn well-generalized representations based on their own data, rather than forcing local models to approach the global model \cite{mendieta2022local}. Moreover, global models differ in their predictions accuracy on different classes, and if the global model confidently and incorrectly predicts samples from certain classes, this inaccuracy may interfere with the training of the local model. Therefore, FedCAD was inspired to propose a class-adaptive self-distillation method that evaluates the degree of trust in the global model's predictions for each category using an auxiliary dataset \cite{he2022class}. This method dynamically controls the impact of distillation based on the performance differences of global model across different classes. Meanwhile, FedICT \cite{wu2023fedict} puts forth a novel strategy by implementing differential learning on client-side models and compensating for the potential loss of global distillation on the server. It makes use of a local knowledge adjustment mechanism to safeguard against skewed global optimization resulting from local updates. Through this mechanism, the local model selectively absorbs distilled knowledge from the global model, effectively mitigating negative impacts without the need for proxy data, modifications to the model framework, or additional information sharing.\\

\noindent\textbf{FD with Local Regularization}. Another approach is using local model to address local model information bias.  More specifically there are,  FEDGEN proposes that clients learn other client knowledge from a lightweight generator which aggregates the output of all local models \cite{zhu2021data}. In order to gain further knowledge of other local models, on the other hand, the FedMMD \cite{yang2022fedmmd} employs a multi-teacher and multi-feature distillation mechanism, introducing teacher middle layer features and logits as distillation targets to gain richer insights from teachers. These allows each client to acquire more abundant knowledge from other participating clients, addressing the effects of information bias on the model. While these methods cope well with client drift, the large amount of information shared increases the risk of information leakage.\\

\noindent\textbf{Summary}. Current algorithms mostly consider reinforcing local models with global model or other participating local models, and few frameworks consider both clients and servers to reduce the feature differences between local and global models. An exemplar in this realm is FedLU, which advocates a mutual distillation strategy that transfers the local model knowledge to the global and absorbs the global model knowledge back \cite{zhu2023heterogeneous}. This strategy has achieved promising results, which can alleviate the excessive reliance on global knowledge, meanwhile avoiding the transfer of knowledge between clients can effectively prevent malicious attacks, worthy of exploring further in depth.

\subsection{FD for Catastrophic Forgetting}
\label{sec4.3}

When training a model using realistic data, the distribution of client data that participate in the training changes over time due to clients continuously collecting data. However, most current FL approaches assume that each client's data remains static, disregarding the dynamic nature of real data. As a result, the global model forgets the knowledge gained from previous tasks when it continues learning new classes~\cite{lee2021preservation}, known as the catastrophic forgetting phenomenon. This forgetting of knowledge presents a bottleneck in the convergence performance of the global model. Catastrophic forgetting primarily arises from the following two factors \cite{ma2022continual}:
(1) Network issues may render clients unavailable during training, causing the data from participating clients to diverge further from the global data distribution.
(2) The local model training process utilizes private data without considering changes in the data distribution, making it prone to overfitting the current distribution of private data.

Catastrophic forgetting can be classified into two primary forms~\cite{liu2023fedet,dong2023no,2020Asynchronous}: (1) \textbf{local forgetting} refers to the loss of knowledge about old classes due to an imbalance between old and new categories; (2) \textbf{global forgetting} refers to the loss of knowledge about the distinctions among local clients resulting from changes in the number of participating clients. The primary cause of catastrophic forgetting is the absence of historical information integrated into the model training process. Motivated by this, FD emerges as a straightforward and effective solution by introducing various distillation mechanisms that facilitate direct learning of historical knowledge or extraction of additional information for knowledge transfer between local-local and local-global models, as outlined in Table~\ref{tb4}. \\

\begin{table}[!ht]
\centering
  \caption{Summarising existing FD algorithms for solving catastrophic forgetting.
}\label{tb4}
  \resizebox{\linewidth}{!}{
  \begin{tabular}{c|c|c|c}
    \hline
    Methods        &Problem type                                &Strategy              &Major tasks               \\ \hline
   CFeD \cite{ma2022continual}  &\begin{tabular}{c} Local forgetting,\\ Global forgetting \end{tabular}  &Continual FD    &\begin{tabular}{c}Text classification,\\ Image classification  \end{tabular}   \\\hline

   FedET \cite{liu2023fedet} &\begin{tabular}{c} Local forgetting,\\ Global forgetting \end{tabular} &\begin{tabular}{c} \begin{tabular}{c} Double distillation,\\ Multiple distillation \end{tabular}  \end{tabular}  &  \begin{tabular}{c}Text classification,\\ Image classification  \end{tabular}\\ \hline

   DFRD \cite{luo2023dfrd} & Global forgetting  & Data-free KD  & Image classification \\\hline
   HePCo \cite{halbe2023hepco}   & \begin{tabular}{c}  Global forgetting \end{tabular}  & latent space KD   &Image classification\\\hline
   Lee et al. \cite{2020Asynchronous}  & Global forgetting  & Clone KD  & Image classification \\\hline

   LGA \cite{dong2023no} &\begin{tabular}{c} Local forgetting \end{tabular}  &Semantic distillation &Image classification \\\hline

   FBL \cite{dong2023federated} & \begin{tabular}{c} Local forgetting\end{tabular}  & Relation distillation  & Image segmentation \\\hline

   TARGET \cite{zhang2023target} & Local forgetting  & KD  & Image classification \\\hline
   FCCL \cite{huang2022learn}  &\begin{tabular}{c} Local forgetting \end{tabular} &Dual-domain KD    & Image classification    \\ \hline
   FedRCIL \cite{psaltis2023fedrcil} & Local forgetting  & Multi-scale KD  & Image classification \\\hline
   STU-KD \cite{zhou2022source}  &Local forgetting  &Collaborative KD    & Image classification    \\\hline
   FedKL-Dist \cite{wei2022knowledge}  &Local forgetting  &Knowledge lock  &Image classification \\\hline
   Liu et al. \cite{liu2022overcoming} &Local forgetting &Local adaptation KD &Image classification \\\hline

   FedSSD \cite{he2022learning}  &Local forgetting &Selective self-distillation  &Image classification \\\hline
   FedLSD \cite{lee2021preservation}  &Local forgetting  &Local self-distillation    & Image classification     \\\hline
   PGCT \cite{liu2023cross}  &Local forgetting  & Prototypical distillation  &Image classification   \\\hline
  FL-IIDS \cite{jin2023fl} & Local forgetting  & self-distillation  & Intrusion detection \\\hline
  PFedSD \cite{jin2022personalized}  &Local forgetting &Self-distillation &Text classification \\\hline
  FLwF-2T \cite{usmanova2021distillation} &Local forgetting &KD &Activity recognition \\ \hline
  \end{tabular} }\\

\end{table}

\noindent\textbf{FD with Historical Data Generator.} With the above analysis, model can effectively mitigate catastrophic forgetting by retraining the historical data. However, sharing a small amount of data provides no sufficient performance improvement, while extensive data sharing may violate the ground rules of FL. The requirement of data privacy restricts the access to historical data. Therefore using generators to synthesize pseudo data similar to historical data to participate in training is kind of effective method. Liu et al. \cite{liu2022overcoming} uses trained global model as a fixed discriminator to train generator synthetic data is used for local adaptation KD, in which the knowledge of the global model is transferred through the synthetic data. Similarly, HePCo \cite{halbe2023hepco} proposes using a lightweight feed-forward network as a conditional generator to generate pseudo-data in the latent space for fine-tuning the global model, which extracts knowledge from the local model via potential space KD. DFRD \cite{luo2023dfrd} utilizes global model and local model to collaboratively train generator for synthesizing data similar to the original distribution that is used for fine-tuning global model via KD to exacting local model knowledge for avoid global catastrophic forgetting. For TARGET proposes by global model training generator, synthesized old task data sent to the client for local model training, which obtains information from the global model through a distillation mechanism to avoid local forgetting \cite{zhang2023target}.\\

\noindent\textbf{FD with Historical Teacher.} Incorporating previous model as teacher models into the training process enriches the information available for student model learning. The historical models can serve as teacher networks to guide the learning of the student models, thereby preventing forgetting in new tasks \cite{dhar2019learning}. Thus there are three strategies for selecting the instructor model: selecting the global model, selecting the local model and selecting the local and global model.\\

\noindent\textbf{1. FD with Global Historical Teacher.} The first strategy is to select the global historical model as the teacher model to guide the current global model or local model training.
Specifically, FBL adaptive class balancing pseudo-labeling mechanism generates old class pseudo-labels to regulate the local model ,and introduces a supervisory mechanism that selects the best global model as the teacher to train the local model by relational distillation \cite{dong2023federated}. And the LGA \cite{dong2023no} proposes category-balanced gradient compensation mechanism and category gradient-induced semantic distillation strategy to train local models, in which the proxy server sends the optimal old global model to the client as a teacher for guiding the training of local models via semantic distillation.\\

\begin{figure}[!t]
\centering
  {\includegraphics[width=3.5in]{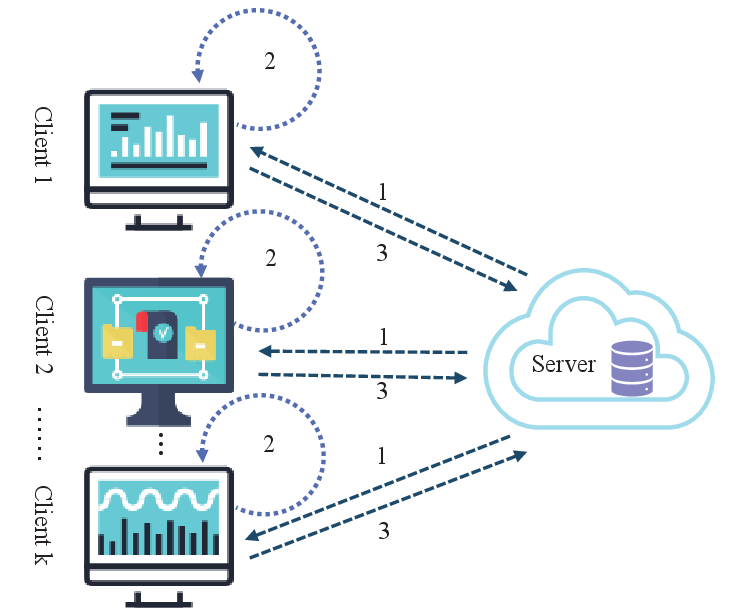}}
  \caption{FD framework with self-distillation mechanism, where 1 means that the local model downloads the initial information of the model from the server, 2 denotes that the local model performs self-distillation training, and 3 represents that the client transmits the information of the trained local model to the server.}
  \label{fig12}
\end{figure}

\noindent\textbf{2. FD with Local Historical Teacher.} The second strategy is to select the local history model as the teacher model to guide the current local model training. Lee et al. \cite{2020Asynchronous} proposes a clonal distillation mechanism to aggregate the local models currently involved in communication into a temporary memory, which is used as an instructor to assist in guiding the global model in the next round of training. FCCL copes with catastrophic forgetting, which utilizes KD for local updating to extract other participants local model knowledge from previous rounds to avoid model updating to lose old class information, while old class distribution is constrained by pre-trained local models \cite{huang2022learn}. Self-distillation mechanisms align well with FL frameworks, guiding local models to assimilate their own past training information. This concept is depicted in Fig.~\ref{fig12}.
For instance, FedLSD introduces local self-distillation within the FL framework, leveraging global knowledge from available data on local devices \cite{lee2021preservation}.
FedSSD  proposes selective self-distillation FL algorithm where the local model learns knowledge from local data while retaining global knowledge to overcome the forgetting problem in the local training phase \cite{he2022learning}. Similarly,
PFedSD introduces a self-distillation mechanism in the FL framework that empowers clients to infuse past personalized model knowledge into the present local model \cite{jin2022personalized}. While FL-IIDS proposes a label-smoothing self-distillation strategy that fuses old class knowledge of previous local model into the labels of the current training data to improve the robustness of the model to each old class, and when the client encounters a new class initializes the local model by the optimal global model \cite{jin2023fl}. These approach ensures the retention of previously acquired knowledge while assimilating new data.\\

\noindent\textbf{3. FD with Hybrid Historical Teacher.} Furthermore, the third strategy is to select the local and global history model as the teacher model to guide the model training. Approaches such as FLwF-2T \cite{usmanova2021distillation} utilize KD-based two teachers strategies to tackle catastrophic forgetting in federated continuous learning. The client's previous model acts as the initial teacher, enhancing the specificity of the student model to proficiently handle prior tasks. Simultaneously, the server serves as a secondary teacher, enhancing the general features of the local model by transmitting the collective knowledge of all participating clients in the training. FedET \cite{liu2023fedet}, on the other hand, proposes a double distillation mechanism applied to the local model and entropy-aware multiple distillation applied to the global model, prompting them to acquire the old class knowledge from the corresponding history model respectively. In the CFeD framework leverages KD in two ways, where the client transfers the knowledge of the old task from the model converged on the previous task to the new model through its own proxy dataset to mitigate local forgetting. The server extracts the knowledge learned in the current and the last rounds into new converged models on independent proxy dataset to fine-tune the converged models in each round to mitigate global forgetting \cite{ma2022continual}.\\

\noindent\textbf{FD with More Knowledge.} There are also other approaches aiming to get more information about the data for the model to mitigate catastrophic forgetting. For one thing, the local models gain more knowledge to mitigate forgetting. For instance, FedKL-Dist proposes extra global model output constraints on distillation losses to solidify the knowledge acquired from other participants \cite{wei2022knowledge}. PGCT \cite{liu2023cross} introduces a prototypical distillation mechanism within the FL framework, facilitating cross-training among clients and enhancing the identification of historical features. The global model acquires more knowledge to alleviate forgetting. FedRCIL proposes multi-scale KD to transfer more local model knowledge to the server, combined with contrastive learning techniques to enable the server to absorb new knowledge while retaining the original knowledge \cite{psaltis2023fedrcil}. The direct fusion of lightweight models trained on different data is a novel exploration, as STU-KD utilizes the alternating direction method of multipliers to consolidate the lightweight global model performance by fusing the lightweight model trained by the server using the source data and the lightweight model trained by the client through collaborative distillation on the target data and soft labels \cite{zhou2022source}. However, in reality source data is often stored on the client side and model fusion is prone to additional information loss.\\

\noindent\textbf{Summary.} In light of the above overview, the FD algorithms dealing with catastrophic forgetting can be divided into three categories: (1) one strategy is to train the generator to generate pseudo-samples of old data for the fine-tuning of the FD framework, thus retaining the ability to process old tasks. However, training a superior generator incurs a large computational overhead, and the generator also sustains catastrophic forgetting. (2) Another strategy is to mitigate catastrophic forgetting by reviewing and consolidating historical knowledge via learning from historical models. However, not all forgotten knowledge is beneficial, and acquiring harmful knowledge from historical models can decrease model performance. Therefore, subsequent research should focus on selective learning of beneficial knowledge from historical models. (3) The last strategy is increasing knowledge transfer during model training which strengthens the robustness of the model to handle historical tasks. However, enhancing the transfer of knowledge during model training not only escalates the demands on client hardware devices but also heightens the risk of privacy breaches. Consequently, overcoming catastrophic forgetting problem is still a problem worth exploring. Follow up studies could consider freezing beneficial historical information from being forgotten or selectively extracting important knowledge to strengthen the model.

\subsection{FD for Communication Costs}
\label{sec4.4}

In the conventional FL framework, data privacy is safeguarded through updates to client-side model information,
obviating the need for data sharing.
This necessitates frequent exchanges of client-side and global model information. Nevertheless, employing model parameters as interaction data leads to a noteworthy escalation in communication expenses, particularly when dealing with larger client populations or more expansive models.
While FL undoubtedly boasts its merits, an over-reliance on mobile devices for decentralized training introduces challenges linked to communication costs.

One of the most formidable challenges in the realm of FL lies in the communication bottleneck
engendered by the frequent exchange of training information among participating devices through constrained bandwidth channels. This bottleneck is primarily a consequence of the following factors:
\begin{enumerate}
    \item The utilization of large networks as backbone systems, which often consist of an immense number of parameters, contributes to this issue.
    \item The FL framework usually accommodates a substantial number of clients, which are required to communicate with the central server in order to transfer information.
    \item In many scenarios, there exists a constraint on transmission bandwidth, thereby extending the time required for frequent uploads.
\end{enumerate}

Mainstream FL algorithms primarily hinge on the utilization of either the parameters or gradients of the local model \cite{mcmahan2017learning}.
In the classical FL framework, the periodic exchange of model parameters entails a substantial communication overhead, contingent upon the model's size \cite{li2020federated}.
When mobile devices are connected to wireless networks and competing for limited radio resources,
the utilization of large models could be restricted, consequently acting as a bottleneck in deploying machine learning models.
Consequently, in situations where communication bandwidth is confined or communication is resource-intensive,
FL's effectiveness may be compromised or even rendered infeasible. Hence, the emergence of a fresh framework that effectively models communication cost pressure becomes imperative.
The reduction of communication costs during model uploads and downloads assumes significance as a pivotal concern in this context.

\begin{table}
\centering
  \caption{Summarising existing FD algorithms for solving communication costs. Reducing Communication Information is noted as RCI, and Reducing Communication Rounds is noted as RCR.}\label{tb5}
\resizebox{\linewidth}{!}{
  \begin{tabular}{c|c|c|c}
    \hline
    Methods        &Solutions                               &Technology              &Knowledge type               \\\hline

  FedED \cite{sui2020feded} & RCI   & KD &Logits \\  \hline
  Etiabi et al. \cite{etiabi2022federated} & RCI   & KD &Logits \\  \hline

  DS-FL \cite{itahara2021distillation} & RCI   & KD &Logits \\  \hline
  Chen et al. \cite{chen2023resource}  & RCI   & KD &Logits \\  \hline
  RFA-RFD \cite{wen2020unified} & RCI   & KD &Logits \\  \hline
  CFD \cite{sattler2021cfd}& RCI   & KD  &Logits\\ \hline

  EDFL \cite{zhang2021distributed} &RCI & Ensemble distillation & Logits\\  \hline
  PTSFD \cite{witt2021reward}&RCI  & Compressed KD & Logits\\  \hline
  FAS \cite{liu2022communication}&RCI   & KD meta-algorithm &Logits \\  \hline
  FedDD \cite{yang2023communication}&RCI   & Double distillation &Logits \\  \hline

   FedKD \cite{wu2022communication}  & RCI   &\begin{tabular}[c]{@{}l@{}}Adaptive mutual distillation\end{tabular}  & Gradients vector\\ \hline 
   CFD \cite{sattler2021cfd}& RCI   & Quantization and delta-coding, KD  &Logits\\ \hline
   HFD \cite{ahn2019wireless}   &RCI  & Hybrid KD &Logits and label covariate vector \\ \hline
   Fed2KD \cite{wen2023communication}   & RCI & Two-step KD &Feature maps \\ \hline
   FedMEKT \cite{le2023fedmekt}   &RCI & KD  &Embedding knowledge \\  \hline
   FedDA \cite{wen2022communication} & RCI   &KD    &Label information              \\ \hline
   SSFL \cite{zhao2022semi}   & RCI  & \begin{tabular}{c}  KD\end{tabular} & Hard labels\\  \hline
   Selective-FD \cite{shao2023selective}  & RCI &KD  &Hard labels  \\  \hline
   HeteFed \cite{liu2023hetefed}  & RCI &KD  &Decomposition logits  \\  \hline
   NRFL \cite{mishra2021network} & RCI & KD  & Model parameters \\  \hline
   FedDTM \cite{liu2023medical} & RCI & KD & Model parameters \\  \hline
   FedUKD\cite{kanagavelu2023fedukd}  &RCI &KD  &Model parameters \\  \hline
   FRL \cite{zhou2023digital}  &RCI  &Bi-distillation  &Model parameters\\ \hline
  SynCPFL \cite{yin2023syncpfl} &RCR  & KD   &Model parameters  \\    \hline
  FedBIP \cite{zhang2023fedbip}  &RCR  & KD   &Model parameters  \\    \hline
  FedMLB \cite{kim2022multi} &RCR  & KD   &Model parameters  \\    \hline

  FLESD \cite{shi2021towards}   & RCI and RCR   & Similarity distillation & Similarity matrices \\ \hline
  FedICT \cite{wu2023fedict}  & RCI and RCR  & Bidirectional distillation & Features, logits and labels \\ \hline
  FedAUXfdp \cite{hoech2022fedauxfdp} &RCI and RCR  & One-shot, FD augmentation   &Logits \\    \hline
  FedKT \cite{li2021practical} &RCI and RCR  &  One-shot, Two-tier KD   &Logits  \\    \hline
  RE-FL \cite{desai2023resource} &RCI and RCR  & One-shot, KD  &Logits  \\    \hline
  Guha et al. \cite{guha2019one}  &RCI and RCR  & One-shot, KD  &Logits  \\    \hline
  LEOShot \cite{elmahallawy2023one}  &RCI and RCR  & One-shot, KD  &Logits  \\    \hline
  Park et al.\cite{park2023towards}  &RCI and RCR  & One-shot, KD  &Logits  \\    \hline
  FedISCA \cite{kang2023one}  &RCI and RCR  & One-shot, KD  &Logits  \\    \hline
  Fedkd\cite{gong2022preserving} &RCI and RCR  & One-shot, KD  &Logits  \\    \hline
  FedSPLIT \cite{eren2022fedsplit} &RCI and RCR  & One-shot, KD   &Non-negative matrices  \\    \hline
  FedKEM \cite{nguyen2023enhancing} &RCI and RCR  & KD   &Logits  \\    \hline
  FedDQ \cite{mo2022feddq}  &RCI and RCR  & Quantization and Coding, KD   &Logits  \\    \hline
  DaFKD \cite{wang2023dafkd} &RCI and RCR  &  Domain aware KD   &Partial parameters  \\    \hline
  FEDGEN \cite{zhu2021data} &RCI and RCR &Data-free KD &Logits \\   \hline

  \end{tabular} }

\end{table}

However, the introduction of KD makes the information communication between the local model and the server in the FL framework unconstrained by the model parameters.
Hence, distillation-based FL simply requires the output of the local model prediction function, which generally has a smaller dimension compared to the model parameters \cite{wen2020unified,chen2023resource}.
This characteristic has led to the consideration of the KD framework, which is instrumental in curtailing the volume of information transmitted during communication.
Table~\ref{tb5} summarizes existing KD-based methods aimed at mitigating communication costs by diminishing communication information.\\

\begin{figure}[!ht]
  \centering
  {\includegraphics[width=3.5in]{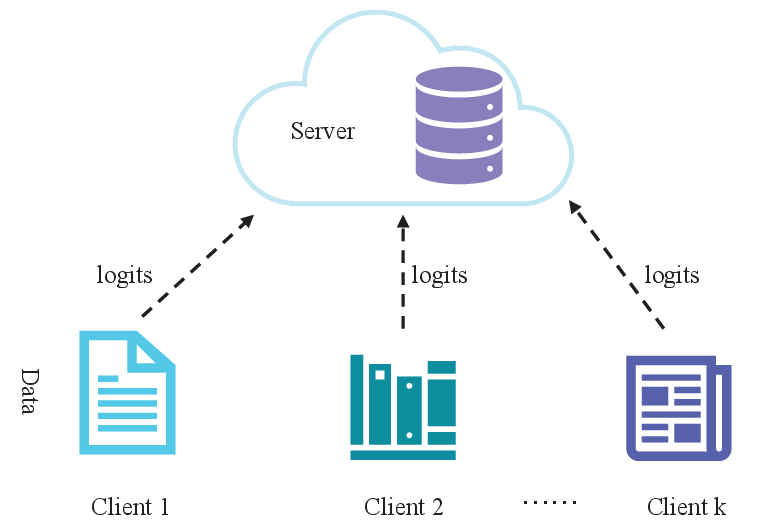}}
  \caption{The local model transfers the output logits information to the server.}
  \label{fig13}
\end{figure}

\noindent\textbf{FD with Reduced Information.} To address the aforementioned issue, the KD mechanism has been introduced within the FL framework to reduce the volume of data in the information exchange process.
A common strategy is to communicate directly between the server and the client using logits from the model output.
In the FD framework, the utilization of KD to transfer logits knowledge instead of the model parameters can effectively reduce communication overhead \cite{zhang2021distributed,sui2020feded,itahara2021distillation,hu2021mhat,etiabi2022federated,liu2022communication}.
For example,
like FedDD proposes a double distillation module to transfer logits information, thereby reducing the volume of information exchanged between local models and the global model \cite{yang2023communication}.
The communication overhead stemming from the model exchange in the FD framework is primarily determined by the output dimension, which is notably smaller than the model parameter size and remains unaffected by the model size.
In these approaches,
clients convey the output logits information of their local models to the server, which then performs aggregation operations to prepare for the next round of training, as depicted in Fig.~\ref{fig13}.\\

\noindent\textbf{FD with Alternative Information}. Additionally, some approaches within the FD framework opt to alter the type of knowledge transferred between models, thus reducing data size in communication \cite{ahn2019wireless,wen2023communication}. An approach like Selective-FD even employs hard label transfer knowledge to further curtail communication costs \cite{shao2023selective}.
Similarly, SSFL mitigates communication costs by minimizing message passing during model training. In this setup, clients convert soft labels into hard labels, upload them to the server, and the server aggregates information using a voting mechanism before transmitting it back to the clients \cite{zhao2022semi}. For FedMEKT proposes using the hidden layer knowledge of the model to enable knowledge transfer between servers and multimodal clients \cite{le2023fedmekt}. And FedDA \cite{wen2022communication} proposes to design knowledge containing labeling information for communication between models using hidden layer based features. These methods are involved in the transfer of much less information than the model parameters and are independent of the model size.\\

\noindent\textbf{FD with Compressed Information.} Some recent methods consider to reduce the amount of information transmitted through the selection of positive information. In such cases, HeteFed proposes binary low-rank matrix decomposition to compress output generated by local models, thus effectively decreasing the amount of information exchanged during communication \cite{liu2023hetefed}. CFD \cite{sattler2021cfd} proposes new quantization and delta coding mechanisms to compress the information of logits participating in the communication. These algorithms essentially revolve around the extraction of essential knowledge by reprocessing information for participation in the communication.
A similar strategy to the above is to directly compress the transmitted information to reduce it. For instance,
 PTSFD introduces a novel reward-based framwork that compresses information to 1-bit quantized soft tags for information transfer in an FD mechanism \cite{witt2021reward}. \\

\noindent\textbf{FD with Reduced Model.} Another approach involves utilizing tiny-size model instead of larger ones for communication with the server. For instance, NRFL \cite{mishra2021network}, FedUKD \cite{kanagavelu2023fedukd} and FedDTM \cite{liu2023medical} reduce the number of parameters communicated between the server and the client through the KD compression local model. For FRL \cite{zhou2023digital} proposes model pruning and federated bi-distillation mechanisms to construct lightweight local models, which reduces uploaded parameters and effectively saves communication time. Similarly, FedKD, in order to further reduce the communication cost, not only proposes an adaptive mutual distillation mechanism to train lightweight models to replace the local model and global model communication, but also introduces a dynamic gradient approximation mechanism for compressing the upload gradient based on singular value decomposition \cite{wu2022communication}.\\

\noindent\textbf{FD with Reduced Rounds.} The previous analysis shows that reducing the number of communication rounds can also effectively mitigate the communication cost of the model. One idea in the FD framework is that the reduction in the number of communications is accompanied by an increase in the speed of model convergence, resulting in an efficient model. For instance, FedBIP \cite{zhang2023fedbip} and FedMLB \cite{kim2022multi} use KD in local model training to accelerate global model convergence to reduce the number of communications. Another idea is to change the information interaction mechanism. A current idea is to categorize clients to reduce the communication between the client and the service period. Like SynCPFL introduces clustering algorithm using GAN and KD, in which the server and the client need to interact only once, whereas other methods often require multiple rounds of interaction \cite{yin2023syncpfl}. A reduction in the number of communications often requires the addition of better mechanism to ensure the diversity of knowledge conveyed, otherwise the reduction in the cost of communications needs to come at the cost of a reduction in accuracy.\\

\noindent\textbf{FD with Reduced Information and Rounds.} A promising way to reduce communication overhead is to reduce the number of communication rounds while reducing the communication information, as presented in Table~\ref{tb5}.
One approach is to reduce the number of communications by speeding up the convergence of the model through the proposed innovative mechanism while reducing the information transferred \cite{zhu2021data,wang2023dafkd}. For instance,
FedKEM introduces a tiny network to assist transferring information between the server and the client, and optimizing the local model using deep mutual learning accelerates the convergence of the global model \cite{nguyen2023enhancing}. Similarly, FedICT framework communicates between the server and the client by means of features, labels and logits that are much smaller than the model parameters \cite{wu2023fedict}. Additional, FedDQ proposes controlled averaging algorithm for aggregating local model outputs to accelerate global model convergence while employ quantization and coding techniques to compress soft labels reducing further the transmitted information \cite{mo2022feddq}. Furthermore, FLESD proposes to adjust the global model by aggregated similarity distillation using client uploaded similarity matrices \cite{shi2021towards}. And these methods the accelerated convergence of the model reduces the number of communication rounds.

\begin{figure}[!ht]
\centering

  {\includegraphics[width=3.5in]{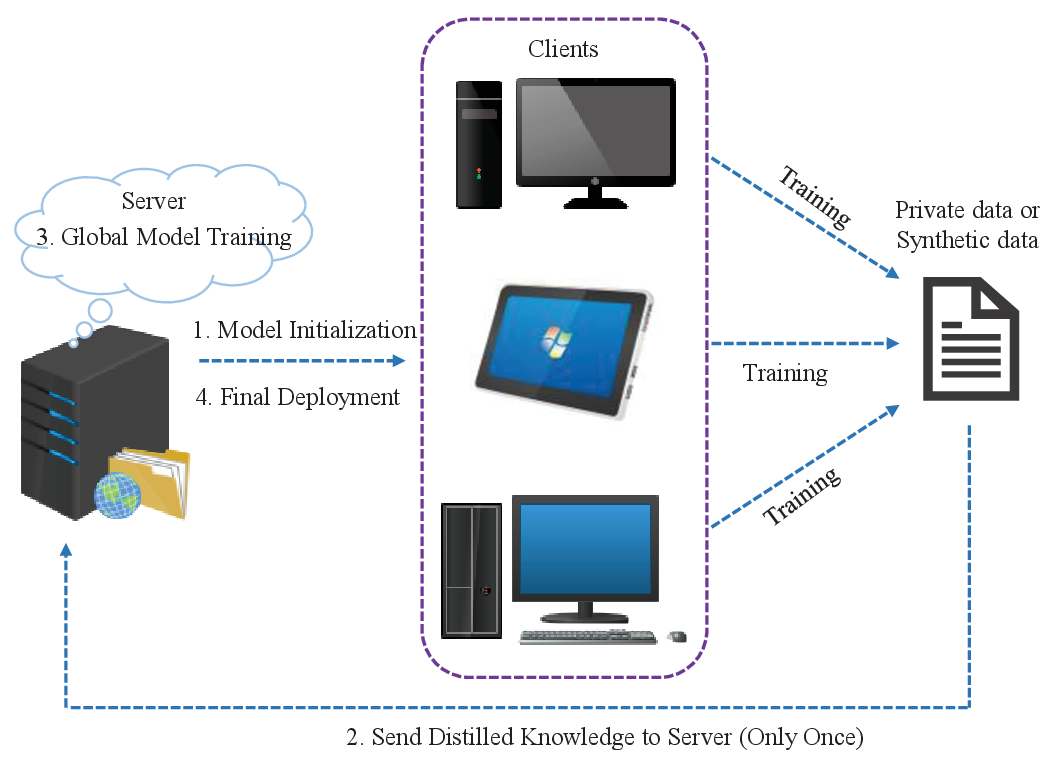}}

  \caption{General structure of the KD-based one-shot FL.}
  \label{fig14}
\end{figure}

Another approach is to use KD to reduce knowledge transfer while introducing a one-shot mechanism to reduce the number of communications.
The one-shot mechanism has been integrated into the FD framework to substantially reduce information exchanges by transferring data in a single round \cite{hoech2022fedauxfdp,li2021practical,gong2022preserving,eren2022fedsplit,guha2019one,elmahallawy2023one,desai2023resource,kang2023one,park2023towards}. The foundational framework structure of one-shot FD is depicted in Fig.~\ref{fig14}.
These algorithms effectively reduce the cost of communication and alleviate bandwidth constraints.
However, it is noteworthy that certain limitations exist within the current one-shot FD approaches. Their applicability is constrained to specific model structures, thereby limiting their broader usage within real-world scenarios.\\

\noindent\textbf{Summary}. To sum up, efforts to reduce communication costs within the FD framework primarily focus on two aspects:
(1) Reducing the number of information transfers and improving the convergence speed of the model;
(2) Reducing the cost of a single information transfer via compressing the model parameters or output.
Many FD algorithms have emerged to enhance communication efficiency, not all methods contribute equally to prediction accuracy. Since these techniques are efficient in alleviating communication pressure, there is a possibility that compressing or randomly selecting transmitted information could neglect crucial data from training.
This underscores the need for continued exploration and refinement of relevant approachs to expand its application scope.
Thus, the development of lightweight model that minimize communication costs while maintaining accuracy remains a crucial research question. However, It's important to acknowledge the trade-off between communication efficiency and prediction accuracy.

\subsection{FD for Privacy}
\label{sec4.5}

FL enables collaborative model building among multiple parties without compromising the confidentiality of their private training data.
This paradigm takes a significant step towards safeguarding data stored on individual devices by exchanging update model parameters instead of raw data.
Nevertheless, the transmission of model parameters throughout the training process raises further concerns about potential privacy and security vulnerabilities \cite{mcmahan2017learning}.
Recent research highlights that the initial FL architecture faces certain privacy and integrity shortcomings.
The utilization of shared model parameters in FL grants the server and other participants access to local model information.
In this context, a malicious actor could exploit model parameters to gain access to pertinent training information.
Moreover, model updates from participants during training might inadvertently expose sensitive data or even enable deep leakage, allowing malicious participants to inject undesired information into the model \cite{melis2019exploiting,zhu2019deep}.
To address these privacy concerns, the incorporation of KD techniques into the FL framework emerges as an effective solution \cite{cheng2021fedgems,wu2023unlearning,gong2021ensemble,gong2022federated}. This approach involves distilling the knowledge from one model to another, thereby facilitating the transfer of valuable insights while minimizing the exposure of sensitive information.
Please refer to Table~\ref{tb6} for a summarization of the various algorithms employed in this context. The section will elaborate on the strategies of privacy protection from the following aspects.\\

\begin{table}[!ht]
\centering
  \caption{Approaches to enhance privacy preservation in FD.}\label{tb6}
  \resizebox{\linewidth}{!}{
  \begin{tabular}{c|c|c|c} \hline
  Algorithms           &strategies               &Influence factor                  & Approaches                \\ \hline

  FedGEMS \cite{cheng2021fedgems}  &\begin{tabular}{c}  Self-distillation,\\  Selective distillation \end{tabular} & Labelled public dataset &Selection and weighting criterion \\ \hline
  FRD \cite{cha2019federated}&Reinforcement KD  &Performance difference &  Policy distillation\\ \hline
  MixFRD \cite{cha2020proxy} &Reinforcement KD  &Performance instability &  Mixup augmented\\ \hline
  FedDefender \cite{park2023feddefender}&Self-distillation  & Synthetic data  & Attack-tolerant global KD \\ \hline
  SPFL \cite{liu2023spfl} &Self-purified  & Computation power  & Self-distillation \\  \hline

  ADFL \cite{zhu2023adfl} &KD  & Fake samples &Adversarial distillation  \\ \hline
  Fedrad \cite{sturluson2021fedrad} &KD &  Number of client &Adaptive distillation \\ \hline
  FedNKD \cite{zhu2022fednkd} &KD & Random noise & Noise data\\ \hline
  Mix2FLD \cite{oh2020mix2fld} &KD & Mixed-up samples &Two-way Mixup algorithm \\ \hline
  KD3A \cite{feng2021kd3a}  &KD  & Domain consensus &Knowledge Vote \\ \hline
  Yang et al. \cite{yang2023fd}& KD &Public data &Membership inference attack\\ \hline

  HeteFed \cite{liu2023hetefed} &KD  &Information loss   & Matrix compression \\ \hline
  Cronus \cite{chang2019cronus} &KD  &Public data & Aggregated scheme\\ \hline

  FedKT \cite{li2021practical}  &Differential privacy &Public data  & Consistent voting \\ \hline
  FedMD-NFDP \cite{sun2020federated} & Differential Privacy &Public dataset  &Argmax distillation approach \\ \hline
  FedHKD \cite{chen2023best} & Differential privacy &Computation power & Transfer hyper-knowledge\\ \hline
  Wang et al. \cite{wang2023can}  & Differential privacy & Public dataset  &Transfer logits \\ \hline
  FedPDD \cite{wan2023fedpdd}  & Differential privacy & Samples limited  & Double distillation \\ \hline
  GFedKRL \cite{ning2023gfedkrl} & Differential privacy & Noise data  & Contrast Learning and KD \\ \hline
  \end{tabular} }
\end{table}

\noindent\textbf{FD with Reduced Communication.} Opting to exchange model logits for information sharing serves as a defence mechanism against gradient attacks. The methods utilizing logits information have been previously summarized, as elaborated in Table~\ref{tb5}, which provides comprehensive details about the applied algorithms. Besides, FedGEMS proposes selection and weighting strategies to reprocess the transferred logits knowledge that avoids malicious and negative knowledge transfer \cite{cheng2021fedgems}. SPFL proposes attention-guided self-distillation techniques that utilize logits and class attention information to purify malicious updates against poisoning attacks \cite{liu2023spfl}. FRD and MixFRD effectively conceals the operational details and access status of the host agent during the transfer of probabilistic information, thus adding an extra layer of security to the communication process \cite{cha2019federated,cha2020proxy}. And Cronus demonstrates that the risk of white-box inference attacks can be eliminated by sharing logits instead of model parameters \cite{chang2019cronus}. Fedrad takes a step further to circumvent the potential influence of malicious clients on the mean value of logits. By leveraging the robust characteristics of medians, Fedrad strategically employs these to partake in the communication of information, thereby enhancing the model's resilience \cite{sturluson2021fedrad}. Analogously, KD3A selects high-quality consensus knowledge transfer based on KD, which enhances the robustness of the model to negative knowledge transfer \cite{feng2021kd3a}.  However, some methods greatly improves data security and privacy from model parameter stealing attacks, the model accuracy has decreased.
Hence, it's important to know that despite these efforts, certain methodologies lack comprehensive privacy guarantees. This becomes a salient issue when proposed attacks don't hinge on specific model structures, gradients, or parameters but rather exploit information leakage from the logits output derived from dataset. This underlines the continuous need for multifaceted strategies that cover both security and performance aspects.

An alternative avenue to bolster the model's robustness against privacy breaches is to reduce the number of model updates exchanged.
This strategic reduction in information exchange not only safeguards the model from privacy leakage attacks but also mitigates the risk of inadvertent information exposure during transmission. The relevant methods have been summarized in Table~\ref{tb5}.\\

\noindent\textbf{FD with Reduced Public Data.} Although, in the pursuit of enhancing model scalability and robustness, the integration of public dataset into the model training process has emerged as a notable strategy. However, it also poses a potential risk to data privacy \cite{takahashi2023breaching}.
This concern becomes even more pronounced in situations where attackers manage to pilfer portions of the local training dataset. Attackers use the difference in weights of client uploaded information to inference and reconstruct localized data leading to sensitive client data leakage.
To mitigate these vulnerabilities, FedNKD synthesizes noisy dataset by adding random noise to real data features. This artificial dataset are then used for training, contributing to enhanced security and privacy during the model learning phase \cite{zhu2022fednkd}. Some other similar methods can be found in the data-free distillation described earlier. Another technique, Mix2FLD takes a unique path to address information leakage concerns by employing a two-way Mixup technique to obscure raw samples. This approach, especially its inverse-Mixup variant, generates synthetic samples for KD, thereby fortifying privacy safeguards \cite{oh2020mix2fld}.\\

\noindent\textbf{FD with Improved Policies.} The privacy protection measures have been integrated into the FD framework, with the integration of differential privacy mechanisms aimed at enhancing the model's information leakage prevention capabilities \cite{hoech2022fedauxfdp,li2021practical,wang2023can,park2023feddefender,ning2023gfedkrl}. For example,
FedPDD \cite{wan2023fedpdd} adopts a double distillation strategy to facilitate local models in acquiring both explicit knowledge about each other and implicit knowledge from their prior training phases. For heightened privacy protection, it combines offline training with differential privacy mechanisms to protect transmitted information.
Further advancements include the noise-free differential privacy mechanism in FedMD-NFDP \cite{sun2020federated} and the Gaussian differential privacy mechanism in FedHKD \cite{chen2023best}. These approaches manage to effectively secure local data privacy with minimal utility trade-offs.\\

\noindent\textbf{Summary}.
Comparatively, FD demonstrates a more potent defense against data privacy breaches than FL. Yet, the potential for user privacy data leakage persists, especially in complex data environments or when subjected to malicious attacks using various methods.
The complexity of data structures or the sophistication of attacks directly correlates with the elevated risk of local data leakage in FD \cite{liu2023mia,yang2023fd,zhu2023adfl}.
It is crucial to acknowledge that as the landscape of data manipulation grows more intricate, the risk of data leakage remains an ever-present challenge in the FD landscape.
Henceforth, data privacy is being protected through three primary approaches:
1) reducing the information transfers;
2) diminishing the reliance on private data engagement;
and 3) incorporating data privacy policies into the model architecture.
With the rise of malicious attack models, future investigations could explore novel model designs employing various combinations of the aforementioned strategies.

\section{Applications}
\label{sec5}

\begin{figure}[!ht]
\centering
  {\includegraphics[width=3.0in]{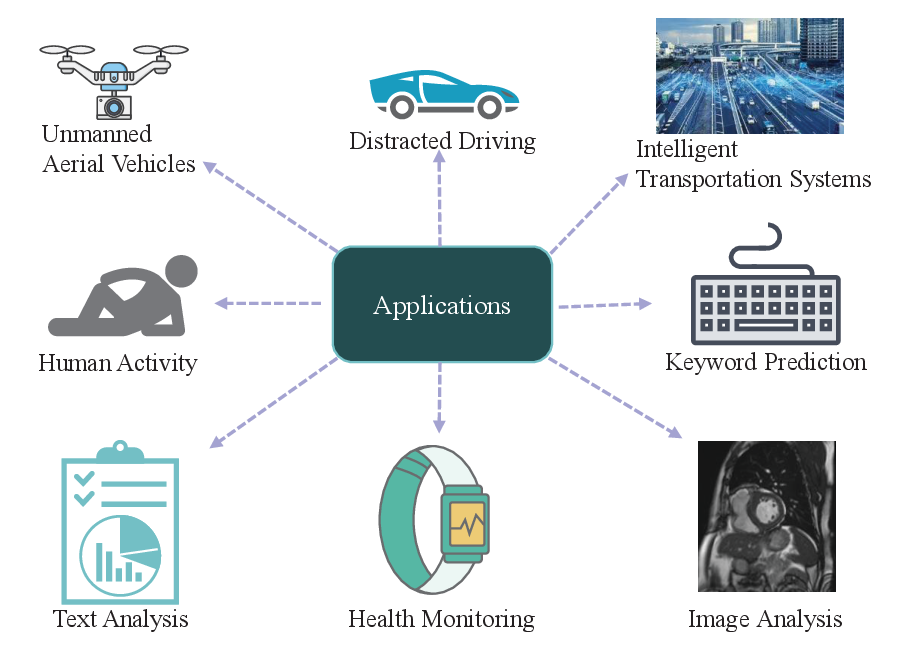}}
  \caption{Utilization of FD in different real-world applications.}
  \label{fig15}
\end{figure}

FD emerges as a more flexible and efficient solution that harnesses the potential of decentralized data across individual clients. Being a novel and potent modeling paradigm, FD offers additional guarantees of privacy and security for both local data and models. Consequently, there has been a series of pertinent investigations aimed at applying FD in practical scenarios, as illustrated in Fig.~\ref{fig15}. In this section, we delve into the applications of FD across various domains, including industrial engineering, computer vision, natural language processing, and healthcare.

\subsection{Application in Industrial Engineering}
\label{sec5.1}

\textbf{Internet-of-Things (IoT)}. The rapid development of wireless communication technology and the Internet has contributed to the growing need for data interaction in the smart industry. The application of FD in industrial engineering is a natural progression, driven by its lightweight nature and robust data privacy protections. With FD, we can leverage the capabilities of IoT's cloud computing and storage to train local models through knowledge transfer. Specifically, HBS proposes bidirectional online KD mechanism to cope with heterogeneous classification tasks in real IoT \cite{wang2020industrial}. And FedBKD proposes a bidirectional KD strategy to mitigate the data and the model heterogeneity that IoT edge devices face in recognizing modulation signals \cite{qi2022fedbkd}. Li et al. \cite{li2023incentive} introduce a KD mechanism to enhance the data compatibility of digital twin-driven based industrial mobile crowdsensing architecture for dynamic sensing of complex IoT environments. FedBIP proposes using KD to accelerate the training process of a wind turbine blade icing prediction model and improve the performance of global model \cite{zhang2023fedbip}.
In terms of energy detection and management, EFCKD proposes an edge-assisted federal comparison KD approach for energy management (detection of energy theft) in urban areas without the need for centralized data collection by electric utilities \cite{zou2023efckd}. For location services, Etiabi et al. \cite{etiabi2022federated} design IoT localization systems suitable for indoor and outdoor based on the proposed FD localization framework.

\textbf{Intelligent Transportation}. In nowadays' fast-paced lifestyle, automobiles have brought great convenience to our daily travels.
However, this surge in the number of vehicles on the road has led to an alarming increase in traffic accidents.
A significant portion of these accidents can be attributed to either distracted drivers or road hazards.
Therefore, there is an immediate need to establish an intelligent transportation system for model training and real-time processing in high-speed moving scenes \cite{zhou2023digital}. For instance, the timely prediction of distracted driving or road obstacles. Specifically, FedBiKD presents a bidirectional KD strategy that facilitates the exchange of knowledge between the server and client, effectively resolving the challenge of data heterogeneity in the context of distracted driving scenarios \cite{shang2023fedbikd}.
In intelligent transportation systems, the FD framework has been introduced to train highly efficient and accurate lightweight model to assist onboard computing devices in monitoring road hazards in real-time \cite{chen2023cooperative}.

Hence, vehicle intelligence and connectivity have gained significant attention. 
They are particularly manifest in the Internet of Vehicles (IoV), where intelligent computing and network communication amalgamate to facilitate traffic management, dynamic information services, and intelligent and safe driving practices.
For instance, EDFL \cite{zhang2021distributed} develope a dynamic map fusion framework in IoV scenarios to obtain high-quality real-time changing maps. And FedDD \cite{yang2023communication} and FedDQ \cite{mo2022feddq} combine the KD mechanism with in-vehicle edge computing to significantly improve data transfer efficiency and mitigate communication link instability caused by vehicle mobility.
In automated vehicle driving, the utilization of visual data information for navigation and decision-making is paramount.
Thus, a confidence-based FD framework has been applied to support automated driving based on visual data, which enhances the reliability of decision-making processes by effectively assessing the confidence levels associated with the received information \cite{chen2023confidence}.
As vehicular technology continues to evolve, these approaches collectively contribute to creating safer, more intelligent, and highly connected road systems.\\

\noindent\textbf{Discussion}. As the IoT continues to evolve, it becomes evident that exploring the realm of system security and stability within this interconnected network of devices is an ongoing necessity.
In the IoT ecosystem, devices are primarily governed by software programs, making the detection of heterogeneous defects in software components a critical requirement. Ensuring the integrity and functionality of these software elements is essential to prevent vulnerabilities that could be exploited by malicious users.
Moreover, the scope of IoT goes beyond software concerns to encompass broader system security and stability. In this context, the detection of system intrusions is crucial \cite{jin2023fl}. By proactively identifying potential unauthorized access attempts or breaches, the IoT environment can be fortified against security threats, thereby protecting the integrity of intelligent systems. The FD scheme is expected to play an increasingly significant role in the industrial application scenarios mentioned above, i.e., requiring significant flexibility to enhance privacy and safety during model training.

\subsection{Application in Computer Vision}
\label{sec5.2}

Despite the extensive research attention devoted to computer vision, there is no conclusive evidence that performance has reached a saturation point, particularly given the ongoing expansion of both model complexity and dataset magnitude. While the emergence of substantial frameworks, exemplified by transformers \cite{touvron2021going,khan2022transformers,liu2023survey}, has undeniably made a significant impact on the field of computer vision, their seamless deployment in real-world settings, notably on resource-constrained edge mobile devices like smartphones and vehicles, remains fraught with challenges.
By harnessing the principles underpinning KD and FL, these models offer a promising avenue to reconcile the pursuit of sophisticated computer vision capabilities with the stringent constraints imposed by real-world edge environments.
Consequently, the exploration of FD models has gained momentum.

\textbf{Image Classification}. Theoretical performance analysis of the FD framework is commonly assessed using benchmark datasets such as Mnist \cite{lecun1998gradient}, Fmnist \cite{xiao2017fashion}, Cifar10 and Cifar100 \cite{krizhevsky2009learning} datasets.
There are numerous works that have been directed towards image classification tasks \cite{hoech2022fedauxfdp,he2020group,he2022learning,cheng2021fedgems,hu2021mhat,chan2021fedhe,cho2023communication,shang2022fedic,wei2022knowledge,chen2023best,sattler2021fedaux,cho2022heterogeneous},
yielding remarkable performance, which can be generalised across multiple domains \cite{huang2022learn,gong2022preserving}.
These FD framework proves the benefits of the FD framework for image classification tasks in computer vision.
Furthermore, FD's application scope has expanded beyond traditional computer vision domains.
For instance, FedUKD has demonstrated utility in monitoring environmental shifts through satellite imagery-based land classification, enabling inference of land use changes in response to climatic and environmental variations \cite{kanagavelu2023fedukd}.

\textbf{Image Recognition}. The FD framework has also been explored in the domain of image recognition.
FedReID improves the accuracy of person re-identification in realistic scenarios by using KD to extract the knowledge generated from local models on a public dataset to optimize the global model \cite{zhuang2020performance}.
Additionally, Liu et al. \cite{liu2022overcoming} proposes an FD-based local adaptive approach that allows the client to have the ability to adapt to new data to overcome forgetfulness and complete the recognition of figures in street photos.\\

\noindent\textbf{Discussion}. Based on the overview provided, several domains within the FD framework still offer untapped potential, signifying considerable room for advancement in computer vision. Specifically, areas such as object detection, semantic segmentation, and video analysis involving small datasets are underexplored and hold substantial avenues for further development.

\subsection{Application in Natural Language Processing}
\label{sec5.3}

\textbf{Keyword Spotting}. Since its initial introduction by Google, the concept of FL has gained significant attention from researchers, particularly in the prediction of users' Gboard inputs on Android devices.
It is combined with distillation strategy has been applied to mitigate false triggers for users and reduce device latency in voice keyword spotting recognition applications \cite{hard2022production}.
With the development of FD, new approaches have emerged to enhance this process.
For instance,
the Fed2Codl model advocates for two-way co-distillation, which has proven efficient and highly accurate in next-character prediction based on textual data \cite{ni2022federated}.
FedLN uses a method of quantifying noise for individual clients in each communication round, thereby refining model performance by rectifying noise-induced samples \cite{tsouvalas2022federated}. It enables accurate keyword predictions based on audio analysis, further demonstrating the practical applications of FD framework.

\textbf{Text Classification}. The FD structure has relevant applications in the field of text classification. Like FedDF and ScaleFL, not only performs sentiment classification but also tackles complex news classification tasks \cite{lin2020ensemble,ilhan2023scalefl}. In scenarios where models need to overcome knowledge disparities among heterogeneous models, the global classifier is trained using unlabeled data from the local model's output.
InclusiveFL introduces an approach for inferential analysis of medical corpora \cite{liu2022no}.
This involves assigning model sizes based on the client's computational capacity and utilizing momentum distillation methods to transfer knowledge from large models to smaller ones.

\textbf{Text Analysis}. The literature has also revealed that FD has found application in a wide range of text analysis tasks, facilitating the extraction of crucial information from textual data.
One notable application is sentiment analysis \cite{cho2022heterogeneous,guha2019one,feng2023adapter}.
DS-FL, for example, presents a novel model for sentiment analysis of online movie reviews.
It employs an innovative entropy reduction aggregation mechanism to manage logits information \cite{itahara2021distillation}.\\

\noindent\textbf{Discussion}. The FD framework can further improve the validity of private language models by fully utilizing accessible datasets without compromising privacy \cite{wang2023can}. Excellent language models that are maliciously attacked to impersonate humans can cost users a great deal of damage. Although FedACK presents a federated adversarial contrastive KD framework that effectively propagates knowledge of data distribution among clients, thus enhancing the response of social bots to real-world issues \cite{yang2023fedack}. With this approach, social bots can be effectively detected and prevented from posing as humans and causing harm to legitimate users. However, while improving the robustness of the model against malicious attacks is still a concern to better enable the application of language models in reality, the deployment of the model on small devices and the requirements of model updates on hardware devices should also be considered.

\subsection{Application in Healthcare}
\label{sec5.4}

The medical domain has witnessed the substantial potential of artificial intelligence technology, particularly in the analysis of extensive medical image datasets.
This technology has demonstrated its capability in early detection of chronic diseases through the examination of medical images.
However, Traditional intelligent healthcare frameworks rely on data collected from individual devices for learning and analysis. Furthermore, in clinical practice, given the sensitive and private nature of medical data, its collection remains challenging, often isolated within distinct medical centers and institutions. This scarcity of data and labeling presents a significant hurdle, leading to the under-performance of tradition machine learning methodologies, thereby impeding the progress of intelligent healthcare applications.
As the healthcare system advances, the urgency to develop distributed intelligence methodologies becomes evident, which aim to realize scalable and privacy-preserving intelligent healthcare applications at the network edge \cite{warnat2021swarm,nguyen2022federated}.
FD has emerged as a promising approach to address these challenges by efficiently training different models while keeping sensitive patient data securely stored on individual devices.

\textbf{Clinical Practice}. FD has demonstrated notable successes in real-world clinical medical images. The benefits brought by FD have been witnessed in the medical fields in classifying medical images and thus identifying the presence of disease \cite{shen2022cd2,wu2023fednoro,kang2023one}.
Specifically, FedCM have derived high accuracy in classification predicting Alzheimer's disease based on mutual distillation \cite{huang2021federated}. And approaches such as FHFL, FedRMD and selective-FD have demonstrated impressive accuracy in discriminating patients with COVID-19 based on lung images \cite{chen2023privacy,shao2023selective}. Additionally, SQMD addresses the challenges of asynchronous training environments and data scarcity in real-world medical contexts, where the validity which had validated on electroencephalogram and electrocardiograph classification data \cite{ye2023heterogeneous}.
The FD framework facilitates the exchange of shared sample knowledge across various institutions, enabling the enhancement of task-specific models through the integration of richer samples information from individual devices. It can harness the collective intelligence of healthcare organizations while respecting privacy considerations. It is noteworthy that FD framework can adeptly manage noise interference in medical image, ensuring reliable performance in the presence of real-world complexities.

The FD architecture has also shown enormous potential for segmenting medical images, in which FedDTM achieves breakthrough segmentation results on the typical COVID-19 dataset and the HAM10000 skin dataset \cite{liu2023medical}. In a segmentation evaluation of 3D magnetic resonance knee images from a private clinical dataset, He et al. \cite{he2023dealing} extensively examined the utility of an FD model based on dual KD. ConDistFL extracts organ and tumor information from partially annotated abdominal CT images \cite{wang2023condistfl}.
The FD not only resolves security issues related to data sharing in remote medical systems but also mitigates the risks posed by client device failures, enhancing the resilience of remote medical systems. This advancement proves invaluable in aiding healthcare professionals in the prevention, diagnosis, and treatment of various diseases within clinical practice.

\textbf{Human Activity Recognition}. In everyday life, the FD framework finds application in the analysis of human activity recognition and the alerting of potentially dangerous behaviors \cite{gudur2020federated,usmanova2021distillation,guo2023fsar,le2023fedmekt}. The monitoring devices play a pivotal role in gathering essential health parameters from individuals.
These parameters are subsequently analyzed using artificial intelligence techniques, thereby facilitating the generation of remote alerts regarding human health status. Specifically,
The FD-based solutions of FL-EDCN \cite{khoa2021fed} introduce personalized cloud-based monitoring frameworks designed for the surveillance of individuals living independently in their homes. This system excels in generating alerts in the event of falls or sudden illnesses, thereby preventing accidents. Furthermore, Jain et al. \cite{jain2021federated} introduce a teacher's assistant mechanism within the FD framework to improve the recognition accuracy for human action recognition in videos. The asynchronous updates adopted by this framework extend its applicability to various sectors, including factory, government, and defence applications.\\

\noindent\textbf{Discussion}. Due to stringent privacy regulations, the healthcare sector faces limitations in aggregating private data to construct public datasets. Consequently, FD models are poised to play a pivotal role in health management, disease screening, diagnosis, and treatment. Serving as an emergent paradigm for deep learning models, FD empowers collaborative training of streamlined models using individual private data, thereby robustly protecting data privacy and security. However, the current FD framework has limited clinical generalizability, owing to the medical image features of different tissues of the human body vary widely and the quantization parameters of different organs are different.

\section{Conclusion}
\label{sec6}

This comprehensive survey have delved into the realm of FD models.
Commencing with an elucidation of the fundamental tenets and advantages of FL and KD,
we subsequently elaborated the pipeline of FD, an emerging paradigm that promises the acquisition of compact models. The survey had discussed in details the challenges that confront FD and a comprehensive analysis of prior endeavors. These initiatives predominantly centered around the development of FD models tailored to address heterogeneity, communication overheads, and heightened privacy concerns.
Moreover, the discourse encompasses the application landscape of the FD framework.
By meticulously evaluating current FD models, it becomes apparent that they hold substantial promise in terms of preserving data privacy, model compression, and performance augmentation. As such, this framework present innovative solutions to real-world challenges, warranting further in-depth exploration and study.

\end{document}